\documentclass{article}

\PassOptionsToPackage{numbers, sort, compress}{natbib}
\usepackage[main,final]{neurips_2026}
\makeatletter                         \renewcommand{\@noticestring}{}       \makeatother  

\usepackage[utf8]{inputenc} %
\usepackage[T1]{fontenc}    %
\usepackage[hidelinks]{hyperref}       %
\usepackage{url}            %
\usepackage{booktabs}       %
\usepackage{amsfonts}       %
\usepackage{amsmath}
\usepackage{nicefrac}       %
\usepackage{microtype}      %
\usepackage{xcolor}         %
\usepackage[capitalize,nameinlink]{cleveref}
\usepackage{graphicx}
\usepackage{subcaption}
\usepackage{listings}
\usepackage{makecell}
\usepackage{enumitem}
\usepackage{comment}

\usepackage{pgfplots}\pgfplotsset{compat=1.18}
\usepgfplotslibrary{groupplots}

\definecolor{flatTurquoise}{HTML}{1abc9c}
\definecolor{flatGreenSea}{HTML}{16a085}
\definecolor{flatEmerald}{HTML}{2ecc71}
\definecolor{flatNephritis}{HTML}{27ae60}
\definecolor{flatPeterRiver}{HTML}{3498db}
\definecolor{flatBelizeHole}{HTML}{2980b9}
\definecolor{flatAmethyst}{HTML}{9b59b6}
\definecolor{flatWisteria}{HTML}{8e44ad}
\definecolor{flatWetAsphalt}{HTML}{34495e}
\definecolor{flatMidnightBlue}{HTML}{2c3e50}
\definecolor{flatSunFlower}{HTML}{f1c40f}
\definecolor{flatOrange}{HTML}{f39c12}
\definecolor{flatCarrot}{HTML}{e67e22}
\definecolor{flatPumpkin}{HTML}{d35400}
\definecolor{flatAlizarin}{HTML}{e74c3c}
\definecolor{flatPomegranate}{HTML}{c0392b}
\definecolor{flatClouds}{HTML}{ecf0f1}
\definecolor{flatSilver}{HTML}{bdc3c7}
\definecolor{flatConcrete}{HTML}{95a5a6}
\definecolor{flatAsbestos}{HTML}{7f8c8d}

\title{Do multimodal models imagine electric sheep?}

\author{%
      Santhosh Kumar Ramakrishnan\thanks{Equal contribution.} \quad             
      Carl Vondrick\footnotemark[1] \quad                                       
      Raja Giryes\footnotemark[1] \\                                            
      \textbf{Philipp Kr\"ahenb\"uhl \quad Vladlen Koltun} \\                   
      Apple                                                                     
    } 

\begin{document}

\maketitle

\begin{abstract}
Yes. We find that large multimodal models develop mental imagery when solving spatial puzzles, and they do imagine sheep when solving sheep puzzles.
We fine-tune a Qwen3.5 VLM to solve twelve diverse visual reasoning tasks -- including tangram, jigsaw, sokoban, 3D mental rotation, and rush hour -- that require understanding geometry, spatial relationships, and the consequences of actions.
By supervising the model to predict the open-loop sequence of actions to solve a puzzle from an initial state, we show that the model's activations after each action encode meaningful visual information about the intermediate state.
This finding suggests that an imperfect visual world model begins to form as a byproduct of learning to select correct actions, in the absence of any explicit visual supervision.
Building on this observation, we propose two ways to sharpen and use the mental images formed by the model.
We find that integrating as few as sixteen visual tokens per step into the chain of thought improves the average solve rate from 83\% to 89\%, with particularly strong gains on reasoning-heavy tasks such as jigsaw and 3D mental rotation.
\end{abstract}

\begin{figure}[b!]
  \centering
      \includegraphics[width=\linewidth,page=1]{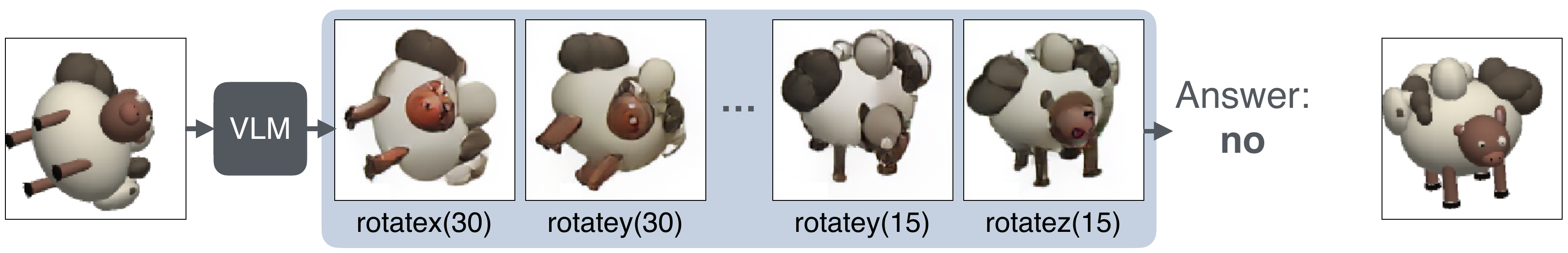}
  \begin{subfigure}{0.14\linewidth}
    \caption{Left sheep}
  \end{subfigure}\hfill
  \begin{subfigure}{0.3\linewidth}
    \caption{Model's imagination}
  \end{subfigure}\hfill
  \begin{subfigure}{0.14\linewidth}
    \caption{Right sheep}
  \end{subfigure}

  \caption{\small \textbf{Are the left and right sheep identical?} We show that VLMs develop mental imagery when trained to solve such spatial puzzles, even in the absence of explicit visual supervision.
  The model takes the left (a) and right (c) images to predict a series of actions that equalize their pose before making a decision on whether the two sheep are identical.
  The center images (b) visualize the model's internal representations as read out by a separate visual decoder.}
  \label{fig:fig_1}
\end{figure}

\section{Introduction}

Mental imagery is a powerful underpinning for spatial cognition, allowing people to visualize and transform objects and environments in their head to answer questions and accomplish goals~\citep{shepard1971mental,ShahMiyake2005,MastJancke2007,GyselinckPazzaglia2012,Nanay2023}.
For example, can you determine whether the sheep in \cref{fig:fig_1} are the same? 
To answer this question, many people will mentally rotate the sheep in their head \citep{shepard1971mental}.

In contrast to humans, frontier models often perform poorly on spatial reasoning tasks, struggling to understand geometric relationships, track objects through transformations, and reason about the effects of actions \citep{ramakrishnan2025does, jia2026omnispatial, chen2026babyvision, li2025elevenplus}.  This gap is not limited to any particular model family or scale, and exposes a fundamental limitation in how current large models represent and reason about visual structure.

\begin{figure}[t]
\centering
    \includegraphics[width=\linewidth]{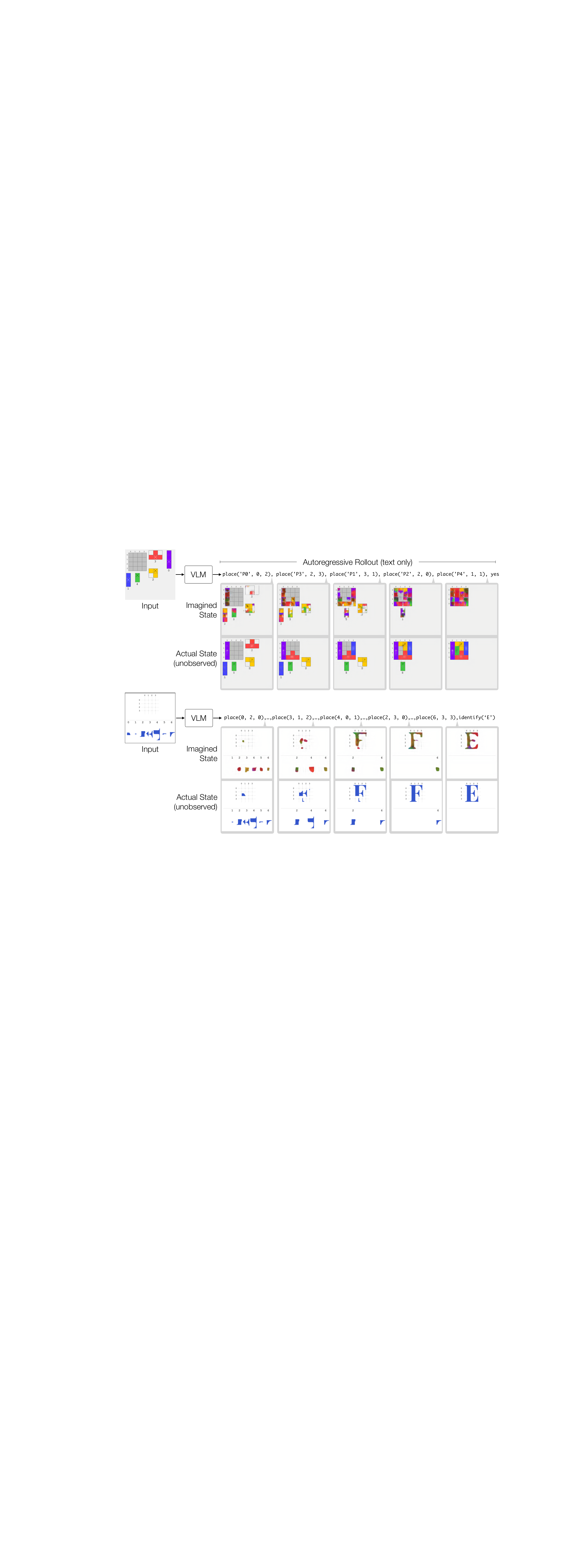}
    \vspace*{-0.2in}
    \caption{\small \textbf{Reconstructing world state from VLM activations.} We freeze a VLM that has been trained to autoregressively predict the actions needed to solve puzzles, attach a visual decoder on the hidden activations at the action boundaries, and train it to predict the visual state. Although the VLM is trained open-loop to predict actions without observing intermediate steps, the decoder is able to reconstruct the state from the VLM's internal representation. The colors are wrong, but the reconstruction of the tangram game matches the geometry well (top), and the identified character emerges in the visual representation (bottom). This suggests that mental imagery emerges in the model.}
    \label{fig:baseline}
\end{figure}

We study to what extent mental imagery emerges in VLMs and how to develop it further for spatial tasks. We use a suite of twelve puzzle classes spanning mental rotation, space filling, jigsaw, and others to analyze the internal representations of multimodal language models.
Working with such puzzles allows us to focus on the core challenge of spatial reasoning in controlled settings. 

We use open-loop behavior cloning, where the model must learn to predict the full action sequence from the first frame without observing intermediate feedback.
Our base model is the recent Qwen3.5 9B \citep{qwen35blog}, which can only solve a small fraction of the puzzles out-of-the-box (13\% average success rate).
Behavior cloning successfully teaches the model to solve some of the puzzles, reaching an 83\% average success rate.
More interestingly, we find that mental imagery emerges in the course of training, even though there is no direct visual supervision. Open-loop behavior cloning creates a supervisory signal that indirectly elicits an internal representation of the evolving world state.
With enough data, mental imagery emerges.
We visualize the model's mental imagery of sheep in \cref{fig:fig_1} and of other objects in \cref{fig:baseline}, showing that the internal representations are granular enough to reconstruct the visual state, recovering object positions, viewpoints, layouts, and spatial relationships that the model was never explicitly asked to represent.
This echoes emergent world representations observed in prior work~\citep{nanda_othello_2023, jin2023emergent,li2022emergent}, now in visually complex environments.

Since multimodal models appear to spontaneously develop mental imagery, we explore how to shape these mechanisms through additional supervision, and how mental imagery can further enhance test-time reasoning.
We develop two approaches. The first adds an auxiliary loss for direct prediction of the visual state after each action.
The second explicitly adds visual tokens into the chain of thought~\citep{wei2022chain}, training the model to autoregressively predict interleaved text and image tokens.

With as few as sixteen visual tokens per action, we see substantial gains on reasoning-heavy tasks such as jigsaw puzzles and 3D mental rotation.
We show an average gain of over 10 percentage points over textual chain-of-thought, especially in tasks such as 3D shape matching or 3D mental rotation, where it is difficult if not impossible to describe the state of the environment in words. Our best approach achieves an 89.5\% average success rate, significantly outperforming the stock base model and other methods.
Our results suggest that shaping and using internal visual representations may advance the spatial cognition of frontier models.

\section{Related Work}

\textbf{Spatial reasoning in vision-language models.}
A large body of recent work has documented the limited spatial cognition of existing vision-language models (VLMs).
Benchmarks such as SPACE~\citep{ramakrishnan2025does}, OmniSpatial~\citep{jia2026omnispatial}, SpatialViz-Bench~\citep{wang2026spatialvizbench}, 11Plus-Bench~\citep{li2025elevenplus}, and Mind the Gap~\citep{stogiannidis2025mindthegap} have demonstrated the limitations of VLMs on mental rotation, spatial visualization, and geometric reasoning.
STARE~\citep{li2025stare} and MIRA~\citep{zhou2025mira} probe multi-step spatial simulation and find that VLMs exhibit inconsistent performance even with oracle visual chain-of-thought guidance during evaluation.
BabyVision shows that the core visual reasoning abilities that develop in humans before language remain out of reach for current models~\citep{chen2026babyvision}, while MOCHI demonstrates large gaps between humans and models on 3D shape inference from multiple viewpoints~\citep{bonnen2024multiview}.
\citet{budny2025visual} trace these failures to deficits in visually-grounded serial processing. \citet{liu2026shellgame} show that VLMs cannot reliably track identical objects through spatial transformations.
\citet{fu2025hidden} find that VLMs substantially underperform their own visual encoders on perception tasks. \citet{asadi2026mirage} expose ``mirage reasoning'' where models produce seemingly correct outputs without grounding in the image.

\textbf{Visual chain-of-thought.}
Several methods aim to augment VLMs with visual reasoning capabilities.
\citet{lotfi2024sketch} propose chain-of-sketch, decomposing visual tasks into intermediate sketch steps for improved out-of-distribution generalization. \citet{menon2024whiteboard} use code generation to allow LLMs to draw before answering.
\citet{bigverdi2024perception} introduce perception tokens -- tokenized visual representations such as depth maps and bounding boxes -- to improve spatial understanding without external tools.
\citet{ray2025mulltokens} propose mull-tokens, modality-agnostic latent tokens for multimodal reasoning.
In the embodied domain, \citet{zhao2025cotvla} show that predicting future image frames as intermediate visual goals improves robotic manipulation.
However, \citet{zeller2026mentisoculi} find with MentisOculi that visual strategies -- from latent tokens to explicitly generated imagery -- generally fail to improve reasoning due to compounding generation errors.
\citet{wu2024mind} propose visualization-of-thought, where LLMs generate intermediate visualizations for spatial reasoning; however, their approach relies on code-based rendering in text-only models rather than learning visual representations end-to-end.
\citet{li2025imagine} engineer explicit visual chain-of-thought by fine-tuning on ground-truth intermediate frames for navigation.
Our work suggests that visual imagination \emph{emerges spontaneously} from just action prediction without any visual supervision, and shows that either explicit compact tokenization (16 tokens per state) or an auxiliary visual loss can sharpen these representations while mitigating drift.

\textbf{Emergent world models.}
A growing body of work shows that sequence models trained via next-token prediction develop rich internal representations without explicit supervision.
\citet{nanda_othello_2023} demonstrates that a GPT trained on Othello gameplay learns a linear, causally relevant board-state representation in its activations.
Gurnee and Tegmark~\cite{gurnee2024space} discover that large language models encode linear representations of space and time at multiple scales, with identifiable spatial neurons.
\citet{jin2023emergent} show that transformers trained on synthetic programs develop representations of unobserved intermediate program states.
\citet{liu2025visualreps} find that visual information is encoded in the internal key-value tokens of multimodal language models, though it often fails to surface in outputs.
We add to this body of work by demonstrating the emergence of visual imagination in multimodal models trained via imitation learning.

\textbf{World models and unified multimodal architectures.}
World models that learn to predict future states from actions~\citep{ha2018world,hafner2023dreamerv3} have seen renewed interest in the context of large generative models.
\citet{maes2026leworldmodel} present LeWorldModel, a joint-embedding predictive architecture that achieves stable end-to-end training from pixels.
\citet{wiedemer2025video} demonstrate emergent zero-shot reasoning in video models including maze solving and physical reasoning. \citet{wang2026demystifying} reveal that reasoning in video diffusion models emerges along denoising steps through a ``chain-of-steps'' process.
\citet{wu2026vega3d} use video generation models to augment multimodal models with geometric cues.
\citet{magne2026nitrogen} train a vision-action foundation model on diverse gameplay via behavior cloning.
Evaluations by \citet{cai2025mmgr} and \citet{guo2025mmecof} show that generative models achieve moderate physical reasoning but fail at abstract spatial tasks. \citet{wang2026vbvr} introduce a large-scale video reasoning benchmark that reveals early signs of emergent generalization.
\citet{heek2026unified} propose unified latent representations regularized by diffusion for images and video.
On the architecture side, unified models that combine visual understanding and generation -- such as BAGEL~\cite{deng2025bagel} and Qwen3.5~\cite{qwen35blog} -- enable both perception and generation within a single model; we use Qwen3.5 as our base architecture and use BAGEL and textual chain-of-thought \cite{MitraCCoT} as baselines.
Unlike approaches that build external world models, we cultivate an internal visual world model via auxiliary prediction during training.

\section{Methods}

We train VLMs to solve spatial puzzles. The model must predict a sequence of actions to solve the game or answer the question in the prompt.

\subsection{Problem Setup}

\begin{figure}[t!]
    \centering
    \includegraphics[width=\linewidth]{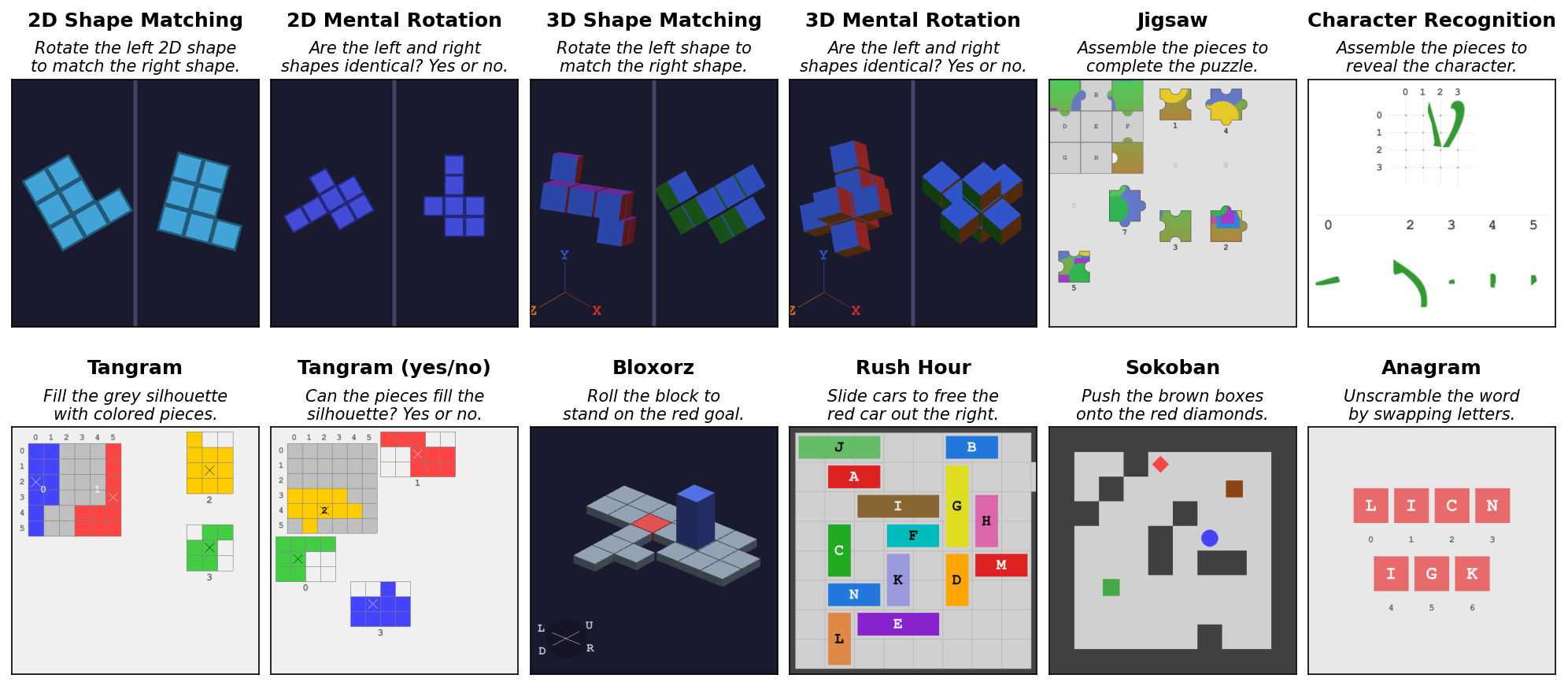}
    \caption{\small \textbf{Spatial reasoning puzzles.} We use twelve puzzles that require a variety of skills such as shape perception, visualization, and planning. We implement optimal solvers to rapidly sample training data on the fly, and use them to study spatial cognition in VLMs.}
    \label{fig:task_gallery}
\end{figure}

Given the initial image $s_0$ corresponding to a puzzle and a text instruction $t$ described in natural language, the VLM $\mathcal{V}$ must produce a sequence of actions that solve the puzzle: 
\begin{equation}
a_1, a_2, \cdots, a_N, r = \mathcal{V}(s_0, t),
\end{equation}
where $a_i$ is the action to be executed at state $s_i$ to solve it. Some puzzles additionally require the model to predict a natural language answer to the puzzle (e.g., yes / no), denoted by $r$. 
The VLM produces the entire action sequence from the initial image alone, without observing intermediate states  that result from its own actions. Since errors cannot be corrected from feedback~\citep{ross2011reduction}, the model must internally simulate the consequences of each action to choose the next one. Open-loop supervision therefore creates pressure to maintain a faithful internal representation of the evolving state, the exact mechanism we hypothesize gives rise to mental imagery.

To study this problem, we implement twelve spatial puzzles shown in \cref{fig:task_gallery}. Our suite comprises well-studied puzzles involving object rotation~\citep{ramakrishnan2025does,bonnen2024multiview,wang2026spatialvizbench}, arranging pieces to form shapes~\cite{lyu2025jigsaw,zhou2025mira,zong2026tangramsr}, sliding objects to reach a target~\cite{van2015pspace,Hauptman2009GPrushUG,yang2021transfer}, and text manipulation~\citep{zhao2023solving,shin2024large}.
Prior work has primarily focused on question-answering versions of these puzzles with small sample sizes, making it hard to perform systematic studies. In contrast, we implement them as game engines (written in Rust for speed) that can rapidly sample a variety of puzzle instances, along with expert solvers (such as breadth-first search, Algorithm X, or closed-form solutions) that can provide supervision via behavior cloning. Please see~\cref{sec:puzzles} for more details.

\subsection{Action Supervision}
\label{sec:no-cot}

\begin{figure}[t]
    \vspace*{-0.1in}
    \includegraphics[width=\linewidth]{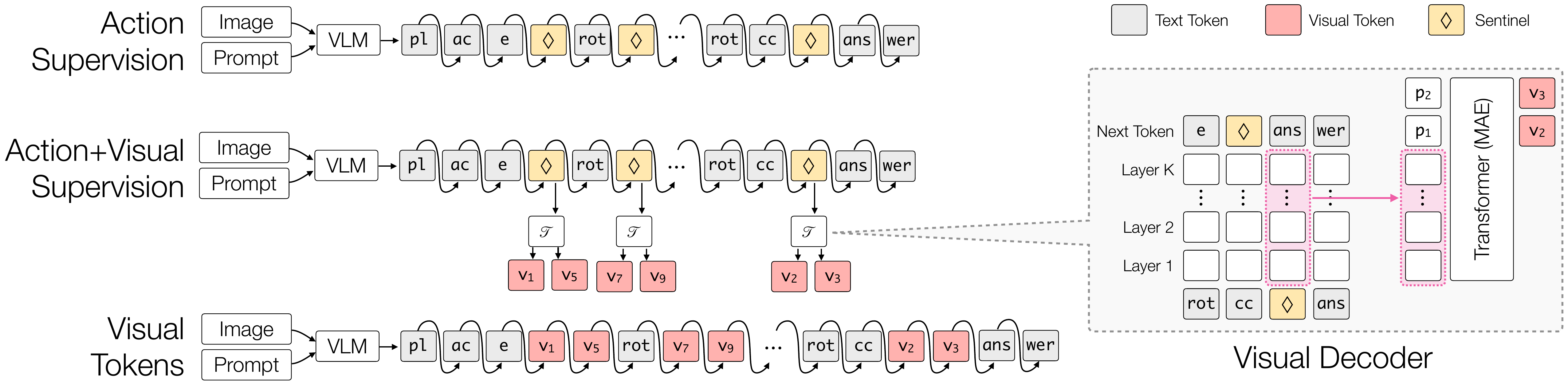}
    \vspace*{-0.1in}
    \caption{\small \textbf{Architectures.} Top: baseline post-trained model, which receives the input image and prompt and directly predicts the sequence of actions needed to solve the puzzle. Middle: visual supervision, which trains a separate head to predict the visual consequence of each action, thereby encouraging mental imagery. Bottom: chain-of-thought with visual tokens.}
    \label{fig:method}
\end{figure}

We train a Qwen3.5 9B model with behavior cloning to predict the sequence of actions needed to reach the goal or answer the question.
The model sees the initial state $s_0$, receives the puzzle instruction $t$, and generates a sequence of actions ($a_1, \cdots, a_N$) supervised by the expert solver (and optionally a response $r$ for puzzles that require a final answer).
The model receives no visual supervision beyond the initial image (see \cref{fig:method} (top) for an illustration). For each puzzle, we run the appropriate expert solver and do full fine-tuning of the base model to predict the next token with teacher forcing.

What makes visual puzzles difficult is that they are sequential: each action changes the world, and the next action must account for that change.
Solving a tangram puzzle with five pieces requires the model to mentally track how each placement and rotation reshapes the available space.
Many of the puzzles require sequences of over twenty steps to solve, and
the state space is inherently visual and geometric.
Pieces have shapes, positions, and orientations, and moves have spatial preconditions and spatial effects. Although the model is trained without any visual supervision (only visual input), the model must learn to maintain an internal representation of the world to plan effectively, and that representation must be rich enough to capture the spatial relationships that determine which actions are legal and which lead toward the goal.

\looseness=-1
The model must learn to represent these states to minimize an open-loop behavior cloning objective.
To probe for emergent mental imagery, we build a visual decoder that is trained separately. We use a four-layer transformer based on masked autoencoders \cite{he2022masked} (see Appendix~\ref{sec:impl_details}). We condition only on the hidden features of the last token of each action, and train to predict the visual state after applying this action. Since the decoder only sees the representation for one token (and not even the initial image), it allows us to understand how well the VLM is tracking the visual state through long-horizons.

\subsection{Visual Supervision}
\label{sec:visual-supervision}

Text-only supervision requires mental images to emerge indirectly. 
By removing the stop gradient on the visual probe and letting the reconstruction loss shape the hidden states, we can strengthen and refine these emergent representations, encouraging the model to form sharper mental images.
During training, we attach the same auxiliary visual prediction head from above that reads the model's hidden states at each action boundary, but now allow the gradients to flow between the two models in order to jointly learn behavior cloning with imagination of the visual state.

The training objective for the visual head combines the standard language modeling objective with the visual prediction loss:
\vspace*{-0.1in}
\begin{align}
    \mathcal{L} = \mathcal{L}_{\text{LM}} + \lambda \, \mathcal{L}_{\text{visual}},
\end{align}
where $\mathcal{L}_{\text{LM}}$ is the next-token prediction cross-entropy over action tokens, $\mathcal{L}_{\text{visual}}$ is the cross-entropy between predicted and ground-truth FSQ codes averaged over all action boundaries in the batch, and $\lambda$ is a weighting hyperparameter.
The gradient of $\mathcal{L}_{\text{visual}}$ flows through the cross-attention back into the language model, encouraging the hidden states at action boundaries to encode the visual state.
At inference time, the visual head is removed and the model generates actions identically to the text-only baseline, with no additional cost.

\subsection{Visual Tokens}
\label{sec:visual-tokens}

We also study visual chain-of-thought, where the model generates discrete visual representations as part of its output sequence.
After each action, the model produces a block of visual tokens enclosed in special delimiters that encode the resulting world state:
\begin{center}
\small
\texttt{<action>} $a_{i_1} \cdots a_{i_M}$ \texttt{</action>}
\texttt{<image>} $v_{i_1} \cdots v_{i_K}$ \texttt{</image>}
\texttt{<action>} $\cdots$ \texttt{</action>} $\cdots$
\end{center}

We resize the rendered $256 \times 256$ image before encoding using the VQ-VAE~\cite{vandenoord2017vqvae} and control the number of tokens per state: resizing to $64 \times 64$ yields a $4 \times 4$ code grid (16 tokens per state), while resizing to $128 \times 128$ yields an $8 \times 8$ grid (64 tokens).
The model's vocabulary is extended with $|\mathcal{C}|$ new tokens, one for each entry in codebook $\mathcal{C}$, and trained jointly with the VLM. The vocabulary embeddings for the new tokens are initialized by aligning the VQ-VAE tokens to the VLM's visual encoder embeddings. Please see the appendix for details.

The training loss decouples visual and text tokens to allow independent weighting:
\begin{align}
    \mathcal{L}_{\text{vcot}} = w \, \operatorname{CE}(H_{\text{vis}}) + (1-w) \, \operatorname{CE}(H_{\text{txt}}),
\end{align}
where $H_{\text{vis}}$ and $H_{\text{txt}}$ partition the supervised token positions into visual-token and text-token subsets, and $w$ is a visual loss weight hyperparameter.
The VQ-VAE encoder and decoder remain frozen; the trainable parameters are the VLM and the new token embeddings.
This approach makes the model's visual reasoning explicit and interpretable: at each step the model must generate an image of the world, which can be decoded and inspected.

\subsection{Implementation Details}
\label{sec:imp_details}

Please see the appendix for full implementation details. The visual    
prediction head and the chain-of-thought both use discrete visual tokens
to represent the world state. We use FSQ~\cite{mentzer2023finite} for the visual supervision head and LlamaGen~\cite{sun2024autoregressive} for the visual chain-of-thought tokens (see ablation in Appendix~\ref{sec:llamagen_vs_fsq}).
The visual prediction head follows an architecture based on masked autoencoders. All methods fine-tune the same model, Qwen3.5 9B~\cite{qwen35blog}. We train each model on one node with 8 B200 GPUs.
Each model trains in 72 hours.
The total compute cost for a single run of all experiments is 9,000 GPU hours.

\section{Experiments}

The goals of our experiments are a) to understand the extent to which mental images emerge in VLMs, b) to assess when they are helpful for visual reasoning, and c) to demonstrate how to sharpen them during learning. We present results across a variety of spatial puzzles, and compare against baselines.

\subsection{Puzzle Suite}
\label{sec:puzzles}

We evaluate on nine puzzle types, three with certificate variants, totaling twelve tasks.
\begin{itemize}[nosep,leftmargin=2em]
\item
\textbf{Spatial rotation~\citep{ramakrishnan2025does,bonnen2024multiview,wang2026spatialvizbench,jia2026omnispatial,zhang2025spinbench}:}
\underline{\smash{3D Shape Matching}} requires rotating a 3D polycube (7--11 cubes) around three axes at $15^\circ$ increments to match a target orientation; \underline{\smash{2D Shape Matching}} is its planar counterpart, requiring rotation of a 2D polyomino into alignment.
\item \textbf{Reconstruction from pieces~\citep{lyu2025jigsaw,zhou2025mira,elkin2025seq2seq,li2025stare,zong2026tangramsr,zeller2026mentisoculi}:}
\underline{\smash{Jigsaw}} requires placing and rotating puzzle pieces ($90^\circ$ CW/CCW) to reconstruct an image on a $2 \times 2$ to $3 \times 3$ grid; \underline{\smash{Tangram}} requires tiling a target silhouette with the seven classical tangram pieces.
\item \textbf{Sliding and pushing~\citep{yang2021transfer,shoham2021solving,taufeeque2025planning,wu2026visual,van2015pspace,Alhassan2019GameOB,Hauptman2009GPrushUG,Cian2022ModelingAS,zeller2026mentisoculi}:}
\underline{\smash{Sokoban}} requires pushing boxes onto goal cells with four-directional moves (boxes can only be pushed, not pulled); \underline{\smash{Bloxorz}} requires rolling a $1 \times 1 \times 2$ block to a goal cell, where some tiles are fragile and break after the block rolls off; \underline{\smash{Rush Hour}} requires sliding vehicles on a $6 \times 6$ grid to free the target car to the exit.
\item \textbf{Text~\citep{zhao2023solving,shin2024large}:}
\underline{\smash{Anagram}} requires unscrambling letters (6--10 characters, at least 5 swaps from the solution) into a valid word; \underline{\smash{Character Recognition}} requires rotating scattered glyph fragments into labeled grid slots and identifying the assembled character.
\end{itemize}
We include certificate variants of tangram and shape matching, where half of the task instantiations are unsolvable and the model must answer `yes' or `no' after optionally trying to construct the solution. For shape matching, we denote the certificate variants as 3D and 2D mental rotation.
Each game is parameterized to control difficulty.
Please see the appendix for the task prompts.

\subsection{Emergence of Mental Imagery}

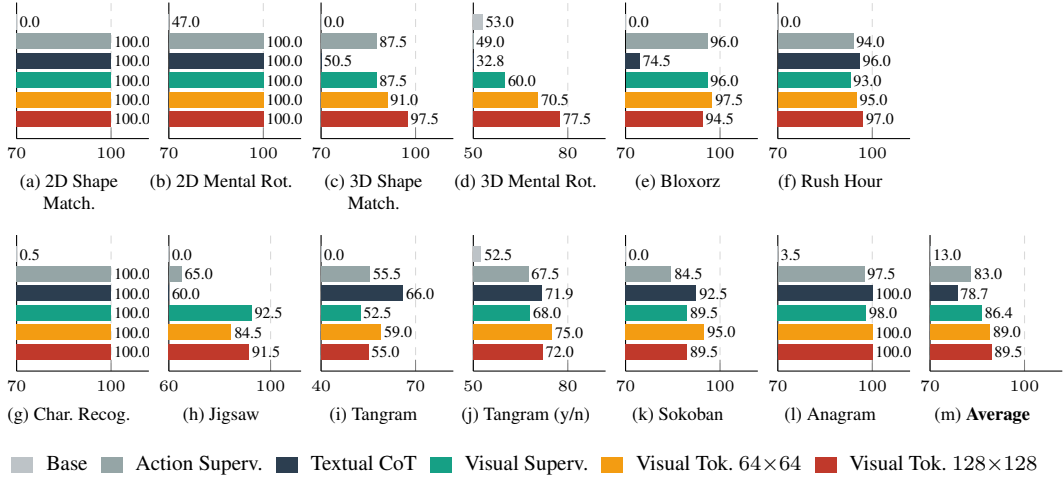
\begin{figure}
\centering
\captionsetup[subfigure]{skip=2pt,font=scriptsize,justification=centering}
\pgfplotsset{
    gameplot/.style={
        scale only axis, width=1.75cm, height=1.8cm,
        ytick=\empty,
        ymin=0, ymax=7,
        xmajorgrids,
        major grid style={dashed, gray!30},
        xticklabel style={font=\tiny},
        every axis/.append style={thin},
        axis x line*=bottom,
        axis y line*=left,
    },
}
\newcommand{\gameplot}[8]{%
\begin{tikzpicture}
\pgfmathtruncatemacro{\xmaxpad}{#8+12}%
\begin{axis}[gameplot, xmin=#7, xmax=\xmaxpad, xtick={#7,#8}]
\pgfmathsetmacro{\va}{max(#1,#7)}
\pgfmathsetmacro{\vb}{max(#2,#7)}
\pgfmathsetmacro{\vc}{max(#3,#7)}
\pgfmathsetmacro{\vd}{max(#4,#7)}
\pgfmathsetmacro{\ve}{max(#5,#7)}
\pgfmathsetmacro{\vf}{max(#6,#7)}
\fill[flatSilver] (axis cs:#7,5.6) rectangle (axis cs:\va,6.4);
\node[right, font=\tiny, inner sep=1pt] at (axis cs:\va,6) {#1};
\fill[flatConcrete] (axis cs:#7,4.6) rectangle (axis cs:\vb,5.4);
\node[right, font=\tiny, inner sep=1pt] at (axis cs:\vb,5) {#2};
\fill[flatMidnightBlue] (axis cs:#7,3.6) rectangle (axis cs:\vc,4.4);
\node[right, font=\tiny, inner sep=1pt] at (axis cs:\vc,4) {#3};
\fill[flatGreenSea] (axis cs:#7,2.6) rectangle (axis cs:\vd,3.4);
\node[right, font=\tiny, inner sep=1pt] at (axis cs:\vd,3) {#4};
\fill[flatOrange] (axis cs:#7,1.6) rectangle (axis cs:\ve,2.4);
\node[right, font=\tiny, inner sep=1pt] at (axis cs:\ve,2) {#5};
\fill[flatPomegranate] (axis cs:#7,0.6) rectangle (axis cs:\vf,1.4);
\node[right, font=\tiny, inner sep=1pt] at (axis cs:\vf,1) {#6};
\end{axis}
\end{tikzpicture}%
}%
\begin{subfigure}[t]{0.135\textwidth}\centering
\gameplot{0.0}{100.0}{100.0}{100.0}{100.0}{100.0}{70}{100}
\caption{2D Shape Match.}
\end{subfigure}\hfill
\begin{subfigure}[t]{0.135\textwidth}\centering
\gameplot{47.0}{100.0}{100.0}{100.0}{100.0}{100.0}{70}{100}
\caption{2D Mental Rot.}
\end{subfigure}\hfill
\begin{subfigure}[t]{0.135\textwidth}\centering
\gameplot{0.0}{87.5}{50.5}{87.5}{91.0}{97.5}{70}{100}
\caption{3D Shape Match.}
\end{subfigure}\hfill
\begin{subfigure}[t]{0.135\textwidth}\centering
\gameplot{53.0}{49.0}{32.8}{60.0}{70.5}{77.5}{50}{80}
\caption{3D Mental Rot.}
\end{subfigure}\hfill
\begin{subfigure}[t]{0.135\textwidth}\centering
\gameplot{0.0}{96.0}{74.5}{96.0}{97.5}{94.5}{70}{100}
\caption{Bloxorz}
\end{subfigure}\hfill
\begin{subfigure}[t]{0.135\textwidth}\centering
\gameplot{0.0}{94.0}{96.0}{93.0}{95.0}{97.0}{70}{100}
\caption{Rush Hour}
\end{subfigure}\hfill
\begin{subfigure}[t]{0.135\textwidth}\
\end{subfigure}\\[0.088in]
\begin{subfigure}[t]{0.135\textwidth}\centering
\gameplot{0.5}{100.0}{100.0}{100.0}{100.0}{100.0}{70}{100}
\caption{Char.\ Recog.}
\end{subfigure}\hfill
\begin{subfigure}[t]{0.135\textwidth}\centering
\gameplot{0.0}{65.0}{60.0}{92.5}{84.5}{91.5}{60}{100}
\caption{Jigsaw}
\end{subfigure}\hfill
\begin{subfigure}[t]{0.135\textwidth}\centering
\gameplot{0.0}{55.5}{66.0}{52.5}{59.0}{55.0}{40}{70}
\caption{Tangram}
\end{subfigure}\hfill
\begin{subfigure}[t]{0.135\textwidth}\centering
\gameplot{52.5}{67.5}{71.9}{68.0}{75.0}{72.0}{50}{80}
\caption{Tangram (y/n)}
\end{subfigure}\hfill
\begin{subfigure}[t]{0.135\textwidth}\centering
\gameplot{0.0}{84.5}{92.5}{89.5}{95.0}{89.5}{70}{100}
\caption{Sokoban}
\end{subfigure}\hfill
\begin{subfigure}[t]{0.135\textwidth}\centering
\gameplot{3.5}{97.5}{100.0}{98.0}{100.0}{100.0}{70}{100}
\caption{Anagram}
\end{subfigure}\hfill
\begin{subfigure}[t]{0.135\textwidth}\centering
\gameplot{13.0}{83.0}{78.7}{86.4}{89.0}{89.5}{70}{100}
\caption{\textbf{Average}}
\end{subfigure}\\[4pt]
\begin{tikzpicture}[
lgsq/.style={minimum width=8pt, minimum height=6pt, inner sep=0pt, yshift=2pt},
lglbl/.style={font=\footnotesize, inner sep=1pt, anchor=base west,
              text height=1.6ex, text depth=.3ex},
]
\matrix[column sep=4pt, nodes={anchor=base}, ampersand replacement=\&]{
\node[lgsq, fill=flatSilver]{};        \& \node[lglbl]{Base};                         \&
\node[lgsq, fill=flatConcrete]{};      \& \node[lglbl]{Action Superv.};               \&
\node[lgsq, fill=flatMidnightBlue]{};  \& \node[lglbl]{Textual CoT};                  \&
\node[lgsq, fill=flatGreenSea]{};      \& \node[lglbl]{Visual Superv.};               \&
\node[lgsq, fill=flatOrange]{};        \& \node[lglbl]{Visual Tok.\ $64{\times}64$};  \&
\node[lgsq, fill=flatPomegranate]{};   \& \node[lglbl]{Visual Tok.\ $128{\times}128$};\\
};
\end{tikzpicture}
\caption{\small \textbf{Solve rates on visual reasoning tasks.} Performance of the stock Qwen3.5 (`Base', without any training on our part), the baseline behavior cloning policy (`Action Supervision'), and various forms of supervison. Chain-of-thought with visual tokens performs best across the puzzle suite.
}
\label{fig:overview}
\end{figure}

We evaluate models on the puzzle suite, comparing their success rates in \cref{fig:overview}. While the off-the-shelf Qwen3.5 model achieves very low performance (13\% average success rate), the base behavior-cloning policy (action supervision, Section~\ref{sec:no-cot}) achieves 83\% average success rate. This is notable and even slightly surprising because the model is trained with open-loop behavior cloning and teacher forcing. During learning, the model needs to plan all the actions to solve the game from a single image, sometimes over a long horizon (20+ steps) -- a situation where errors can compound. What is the model learning in order to achieve this?

\begin{figure}
    \centering
    \includegraphics[width=\linewidth]{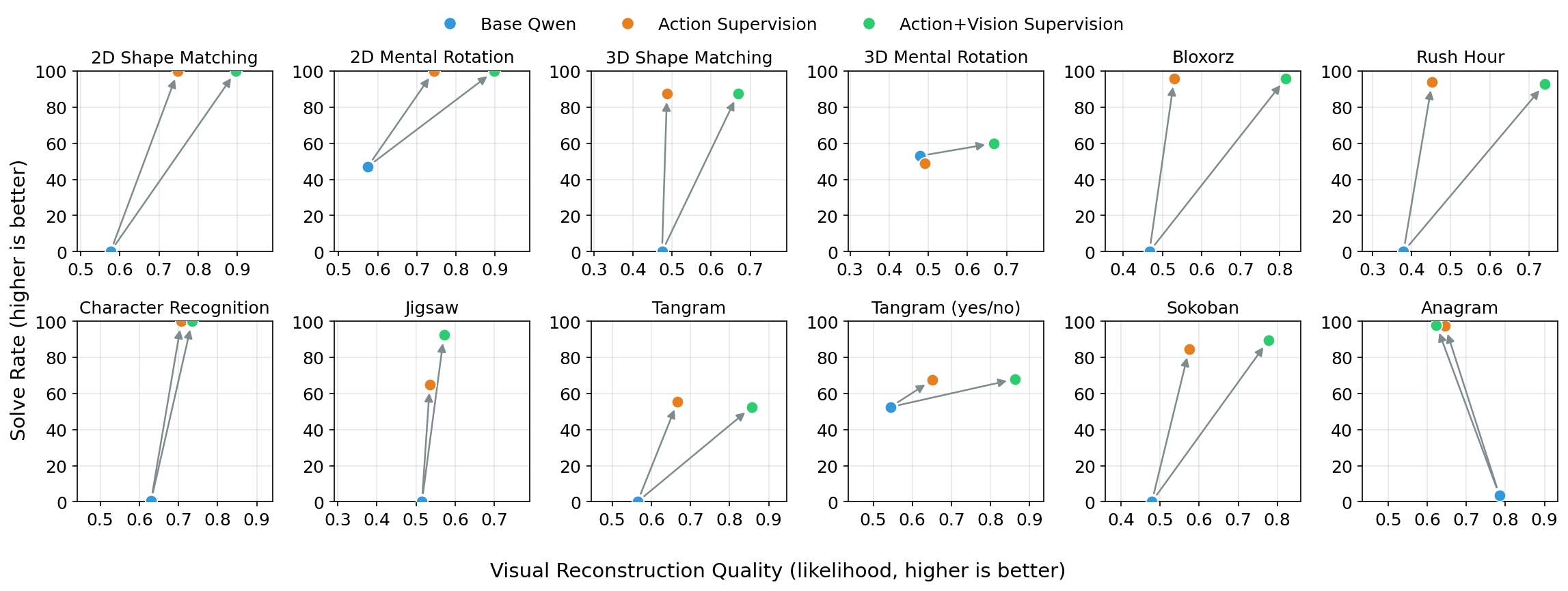}
    \vspace*{-0.2in}
    \caption{\small \textbf{Solve rate versus mental image quality.} We train visual decoders on the base Qwen3.5 9B model (blue) and our post-trained models (orange, green). We plot their success rates (vertical axis) and mental image fidelity (horizontal). When an arrow points toward the top right, higher visual fidelity is accompanied by a higher success rate -- a trend we see for ten out of the twelve games.}
    \label{fig:psnr_vs_solve}
\end{figure}

We train a visual decoder on the frozen behavior-cloned LLM, which allows us to visualize and inspect what is encoded at each step in the autoregressive rollout. Although no gradients flow from the visual head to the LLM, and there is no visual supervision on the LLM, we are able to reconstruct significant detail about the intermediate states during the LLM rollout. \cref{fig:baseline} visualizes two examples, where although the colors are off, the geometry of the tangram grid is reproduced well (top) and the correct shape emerges as the VLM imagines piecing the character together (bottom). Furthermore, we plot the correlation between solve rate and mental image quality in \cref{fig:psnr_vs_solve}. We can see that action supervision elicits higher-fidelity mental imagery.

This raises a question: if the model already encodes partial visual state in its representations, can we improve performance by explicitly encouraging this behavior?
The imperfect quality of the reconstructions suggests that the emergent encoding, while useful, leaves room for sharper internal representations.
\cref{fig:psnr_vs_solve} shows that, for games with initially low solve rate, removing the stop gradient on the visual probe, and thus explicitly supervising the internal representations to be predictive of the visual state, can lead to dramatic gains. This can be seen in the sharp pivot to the right in 3D mental rotation, 3D shape matching, and other puzzles in \cref{fig:psnr_vs_solve}. For most tasks, the model's solve rate improves with the fidelity of the mental images. Anagram is an exception, where the base model has near-zero solve rates but better reconstructions than post-trained models. Upon inspection, we found that the base model produces less informative reconstructions that are quantitatively better since the image contains very few foreground pixels (see Appendix~\ref{sec:anagram} for more detail).

\begin{figure}
    \includegraphics[width=\linewidth]{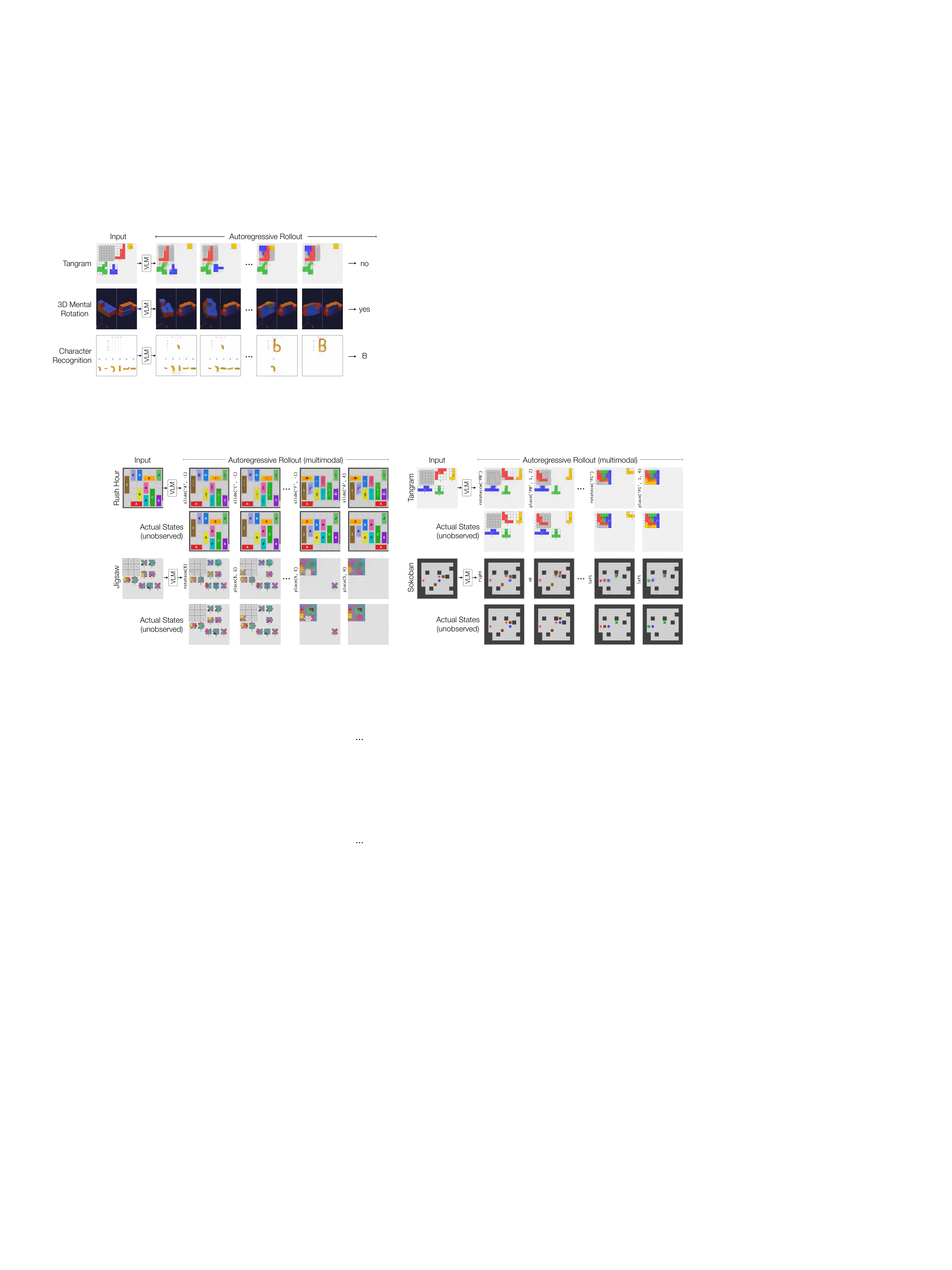}
    \caption{\small \textbf{Visual chain-of-thought (QA).} We show rollouts with explicit visual thinking tokens for QA puzzles, where the model must decide whether a puzzle is solvable or not, or identify an object/character. The model forms mental images to help answer these questions.}
    \label{fig:vcot}
\end{figure}

\begin{figure}[t!]
    \includegraphics[width=\linewidth]{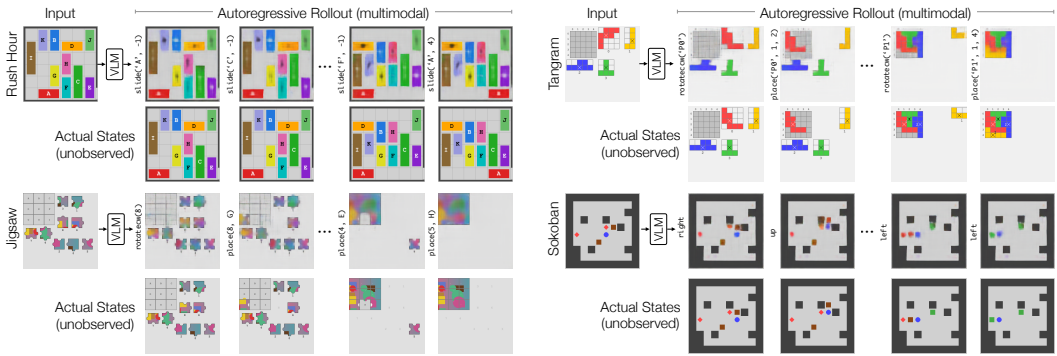}
    \caption{\small \textbf{Visual chain-of-thought (gameplay).} We show the visual thinking tokens generated by the model during gameplay. The VLM imagines the consequences of its actions as it plans its moves, then executes them.}
    \label{fig:vcot_gp}
\end{figure}

\subsection{Game Playing and Question Answering}

We compare six methods that differ in how intermediate state information is incorporated during training. We include the methods introduced previously: action-only supervision from~\cref{sec:no-cot}, visual supervision from~\cref{sec:visual-supervision} that further shapes the internal representations to be predictive of the world state, and two versions of visual tokens from~\cref{sec:visual-tokens} that decode $64\times 64$ or $128 \times 128$ states after each action. Additionally, we include a textual chain-of-thought baseline (based on the same Qwen3.5 9B model) inspired by~\citet{MitraCCoT} that replaces visual tokens with structured JSON scene-graph descriptions (see Appendix~\ref{sec:text_cot_details} for details).
Each game is evaluated on 200 held-out puzzles with greedy autoregressive decoding using vLLM~\cite{kwon2023efficient}.

\begin{figure}
\centering
\includegraphics[width=0.9\linewidth]{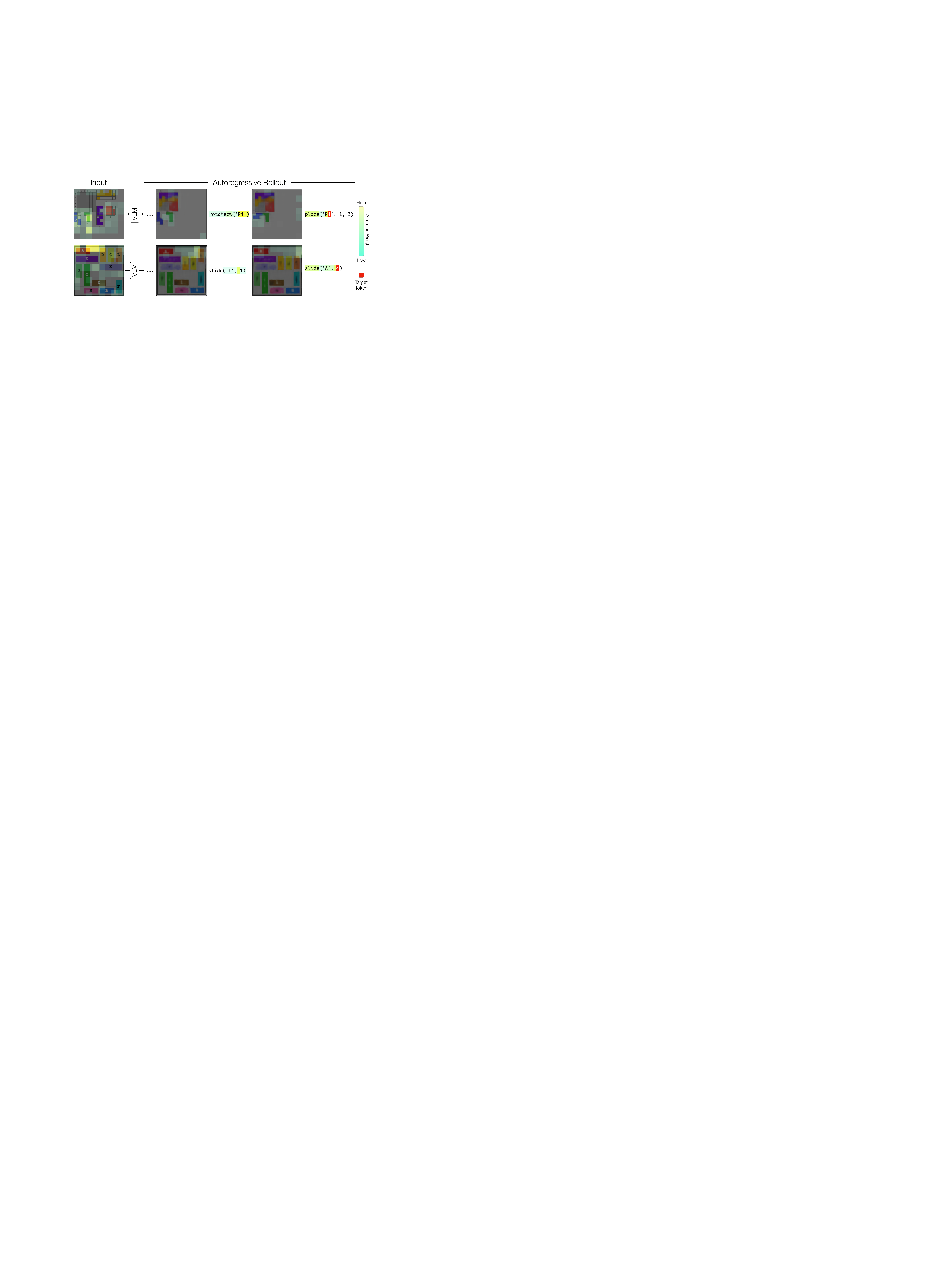}
\caption{\textbf{Attention weight on visual chain-of-thought.} We visualize the attention weights for a target token (in red). The VLM with visual tokens is able to attend to its own mental images to help plan actions. The top row shows the model attending to the rotated piece to decide its location, and the bottom row shows the model attending to the free space to decide how to win the rush hour game.}
    \label{fig:attnvis}
\end{figure}

\textbf{Visual chain-of-thought is effective.}
As shown in \cref{fig:overview}, explicitly generating visual tokens achieves an average solve rate of 89.5\%. It outperforms the textual CoT baseline and visual supervision by 10 and 3 percentage points on average, respectively. See~\cref{fig:vcot,fig:vcot_gp} for illustrations of the visual tokens generated, and~\cref{fig:attnvis} for a qualitative analysis of the attention weights over the generated visual tokens.

\textbf{Visual supervision refines mental imagery.}
Mental imagery emerges naturally without direct visual supervision. Further shaping this ability through visual supervision improves the fidelity of mental images (\cref{fig:psnr_vs_solve}) and improves the average solve rate by 3.4 percentage points (\cref{fig:overview}).

\textbf{Compact tokenization is effective.}
The two resolutions for visual tokens produce complementary per-game profiles.
The $64 \times 64$ variant, with only 16 tokens per state, is better on long-horizon games like sokoban, tangram, and bloxorz, while the $128 \times 128$ variant with 64 tokens per state is better on tasks like jigsaw, 3D shape matching, and 3D mental rotation, where higher resolution for intermediate states proves beneficial.
Despite these differences, their overall scores are close (89.5\% vs.\ 89.0\%), suggesting that even highly compressed visual representations carry enough spatial information to improve action selection on balance.
The compactness of the $64 \times 64$ representation is particularly notable: 16 tokens per state adds minimal sequence length overhead while still capturing sufficient spatial structure to improve planning.

\textbf{Comparison to text-based reasoning.}
Textual CoT \cite{MitraCCoT} uses structured JSON scene graphs as intermediate representations (instead of visual tokens), and generates them autoregressively through behavior cloning.
It achieves an average success rate of 78.7\%, which falls well short of the other methods. This approach struggles on tasks such as 3D shape matching, 3D mental rotation, jigsaw, and bloxorz.
This suggests that the benefit of visual chain-of-thought stems specifically from the visual modality, not just from the expenditure of test-time compute. Structured textual descriptions of the state, even when compact and machine-readable, do not provide the same representational advantage as visual tokens. Additional details are provided in the appendix.

\textbf{Additional studies.}
We evaluated fine-tuning BAGEL~\cite{deng2025bagel}, a unified understanding and generation model, but it did not reach reasonable performance. See the appendix for this and other studies.

\section{Conclusion}

We studied the emergence of mental imagery in vision-language models trained to solve multi-step spatial puzzles.
We showed that models trained with open-loop behavior cloning learn implicit visual representations of the evolving world state. Sharpening mental imagery via additional supervision further improved performance on spatial puzzles with no test-time overhead. Introducing chain-of-thought reasoning that incorporates visual representations increased the model's success even more.
We hope that this work contributes to a better understanding of spatial cognition in frontier models.

\bibliographystyle{unsrtnat}
\bibliography{ref}

\clearpage
\appendix

\section{Implementation Details}
\label{sec:impl_details}

\textbf{Visual tokens.} We use discrete visual tokens to represent images in all of our methods. We experiment with two encodings. Firstly, we use an FSQ autoencoder with $D = 6$ dimensions with $L = 5$ levels each, yielding a codebook of $5^6 = 15{,}625$ entries.
It maps $256 \times 256$ images to a $16 \times 16$ grid of codes via a convolutional encoder with four stride-2 stages. Secondly, we also explore LlamaGen VQ-VAE~\citep{sun2024autoregressive} which performs $16\times$ spatial downsampling and a codebook of 16{,}384 entries. We found that LlamaGen tokens worked best for autoregressive prediction and FSQ worked best for visual supervision. See~\cref{sec:llamagen_vs_fsq} for supporting experiments.

\textbf{Visual prediction head.} We follow an architecture based on masked autoencoders \cite{he2022masked}.
The decoder operates at dimension $d = 256$ with $8$ attention heads.
Hidden states from all $N_{\text{LLM}}$ LLM layers at the visual prediction position are projected from the
LLM hidden dimension $H$ (for Qwen3.5 9B, $H = 4096$ and $N_{\text{LLM}} = 32$) down to $d$, yielding one
memory token per LLM layer that the decoder attends to.
The self-attention stack consists of $4$ pre-norm transformer layers with GELU activations, feed-forward
dimension $4d = 1024$, and no dropout.
The code prediction head is a single linear layer mapping $d \to D \times L = 30$ logits per patch. In total, the visual head has approximately 36.9M trainable parameters for the 9B model. The frozen FSQ autoencoder adds no trainable parameters.

\textbf{Base model.}
All methods fine-tune the same base model, Qwen3.5 9B~\cite{qwen35blog}, which is cast to \texttt{bfloat16} with FlashAttention-2.
Training uses FSDP with full sharding and activation checkpointing.
Sequences are packed to a maximum of 1{,}024 tokens. 

\textbf{Optimization.}
All methods use AdamW \cite{loshchilov2017decoupled} with $\beta = (0.9, 0.999)$, weight decay of $0.01$, gradient clipping at $1.0$, and a cosine learning rate schedule with 1{,}000-step linear warmup and minimum LR ratio $0.1$, for up to 25{,}000 iterations.
The base model and visual head variants use a learning rate of $1 \times 10^{-5}$; the visual token variants use $3 \times 10^{-5}$ to accommodate learning the additional token embeddings.
For the visual head, its parameters use a $100\times$ learning rate multiplier.
When an episode contains more than 32 action boundaries, we randomly subsample visual frames to bound the per-step cost of the visual head. %

\begin{figure}
\centering
    \includegraphics[width=0.9\linewidth]{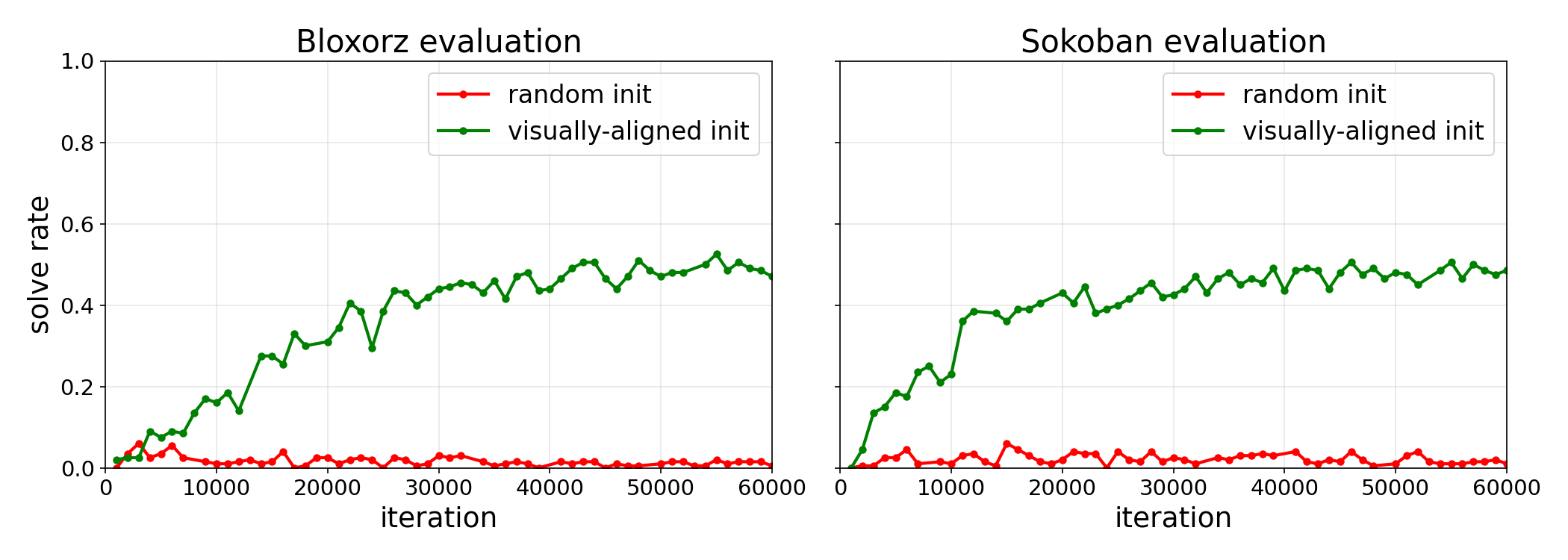}
    \caption{\textbf{Embedding initialization for visual tokens.} We compare the solve rates versus training iterations for two embedding initialization schemes when visual tokens are added to the model's vocabulary. When the embeddings for the tokens are randomly initialized, the model fails to train on a representative set of sokoban and bloxorz games. After our proposed initialization by aligning the VQ-VAE's visual tokens with the Qwen3.5 VLM's visual encoder embeddings, the model starts to learn puzzle solving rapidly.}
    \label{fig:embed_init}
\end{figure}

\textbf{Embedding initialization for visual tokens.} As discussed in~\cref{sec:visual-tokens}, we add $|\mathcal{C}| = 16,384$ VQ-VAE tokens from LlamaGen to the model vocabulary. We observed that initializing the vocabulary embeddings to Gaussian noise resulted in slow convergence, and would result in the model being stuck at nearly 0\% solve rates on most games. We identified the root cause of this was the visual tokens in the vocabulary were disconnected from the model's visual embedding space. Specifically, whenever the model encountered visual tokens in its context, it would not be represented in the same way as an image patch would by the visual encoder, but rather as a learnable and randomly initialized embedding. To fix this, we collected a dataset of images sampled from the puzzles via the expert solver. On each image, we obtained the discrete token representation using the VQ-VAE as well as the visual embeddings from the Qwen3.5 visual encoder. Since the tokens and the embeddings are spatially aligned, we can associate Qwen3.5 visual embeddings to corresponding VQ-VAE tokens. We averaged the visual embeddings corresponding to each VQ-VAE token over the collection of images, and used the average visual embedding per token to initialize the corresponding entry in the vocabulary embedding matrix. This ensured that whenever the Qwen3.5 LLM encountered a visual token, it would be similar to the visual embeddings it would see from an image, and made a massive difference to learning. See~\cref{fig:embed_init} for a comparison.

\textbf{Text instructions for puzzles.} We display the text prompts $t$ for each puzzle (see \cref{sec:puzzles}) in~\cref{tab:task_prompts}.

\begin{table}[t]
\centering
\small
\renewcommand{\arraystretch}{1.3}
\begin{tabular}{@{}p{0.22\textwidth} p{0.72\textwidth}@{}}
\toprule
\textbf{Game} & \textbf{Task Prompt} \\
\midrule
Tangram & Complete the tangram. Fill the grey silhouette using the colored pieces. \\
Tangram (certificate) & Determine if the pieces can fill the silhouette. Say `yes' if they can, `no'
if they can't. \\
Shape Matching (3D) & Rotate the left shape to match the right shape. \\
Mental Rotation (3D) & Determine if the left and right shapes are the same. Rotate to
check, then say `yes' if they match or `no' if they don't. \\
Shape Matching (2D) & Rotate the left 2D shape to match the right shape. \\
Mental Rotation (2D) & Determine if the left and right 2D shapes are the same (possibly
rotated). Rotate to check, then say `yes' if they match or `no' if they don't. \\
Jigsaw & Reconstruct the original image by rotating each piece to the correct orientation and placing
it at its correct grid position. \\
Anagram & The letters of a word have been scrambled. Swap letters to unscramble the word, then submit
with \texttt{identify(word)}. \\
Character Recognition & A character has been broken into pieces that are scattered and rotated. Rotate
each numbered piece to the correct orientation and place it into the matching lettered slot in the
assembly area. Once all pieces are correctly assembled, identify the character. \\
Sokoban & Solve this Sokoban puzzle. Push the brown boxes onto the red diamonds. \\
Bloxorz & Solve this Bloxorz puzzle. Roll the block so it stands on the red
goal.\textsuperscript{$\dagger$} \\
Rush Hour & Slide vehicles to free the red car (A) and move it to the right edge exit. \\
\bottomrule
\end{tabular}
\vspace*{0.1in}

\caption{\textbf{Puzzle task prompts.} Task prompts shown to the model for each game in our puzzle suite. \textsuperscript{$\dagger$}~When fragile tiles are present, the prompt appends: ``Striped tiles are fragile and will break after the block leaves them.''}
\label{tab:task_prompts}
\end{table}

\clearpage
\section{Textual Chain-of-Thought Baseline Details}
\label{sec:text_cot_details}

We describe the textual chain-of-thought (Textual CoT) baseline in detail.
Inspired by the compositional scene-graph prompting framework of \citet{MitraCCoT}, we replace the visual token channel with a structured JSON description of the state emitted between \texttt{<image>} and \texttt{</image>} delimiters at each action boundary.
The training pipeline is identical to the visual chain-of-thought variants: the model is teacher-forced to reproduce the ground-truth scene-graph JSON at every step, using the decoupled loss with weight $w = 0.3$.
The only difference between textual CoT and VQ training is what sits between the \texttt{<image>} delimiters -- JSON text versus discrete visual codebook indices.

We design a per-game JSON schema that encodes the minimum information needed to reconstruct the state plus a small number of cheaply derivable quantities (counts, distances, flags).
The schemas are calibrated to produce a max of 64 tokens per state when tokenized with the Qwen3.5 tokenizer, matching the visual chain-of-thought with the larger image of size $128\times 128$. We tried using longer textual chain-of-thought but it did not work better than the current setup. For some games, fewer tokens sufficed given the simpler state description. 
\cref{tab:text_cot_tokens} reports the measured token statistics across all game types, computed over 100 episodes per game along optimal solution paths.
The QA variants use the same schema as their base game and have nearly identical token statistics.

\begin{table}[h]
\centering
\small
\caption{\textbf{Textual CoT token counts per state description.} Statistics measured over 100 episodes per game along optimal solution paths, tokenized with the Qwen3.5 tokenizer.}
\label{tab:text_cot_tokens}
\begin{tabular}{lrrrr}
\toprule
Game & Mean & Median & Min & Max \\
\midrule
3D Shape Matching / Mental Rotation & 64.7 & 65 & 58 & 71 \\
2D Shape Matching / Mental Rotation & 31.0 & 31 & 29 & 33 \\
Jigsaw & 59.5 & 65 & 35 & 80 \\
Tangram & 47.5 & 49 & 28 & 68 \\
Character Recognition & 39.8 & 39 & 34 & 49 \\
Sokoban & 42.7 & 41 & 27 & 64 \\
Bloxorz & 48.1 & 48 & 38 & 61 \\
Rush Hour & 54.7 & 54 & 44 & 68 \\
Anagram & 31.9 & 32 & 28 & 37 \\
\bottomrule
\end{tabular}
\end{table}

All schemas include a \texttt{moves} field recording the number of actions taken so far.
Below we describe the schema and provide a concrete example for each game type.

\textbf{Sokoban.}
The grid is encoded as a string with characters \texttt{\#}~(wall), \texttt{.}~(floor), \texttt{@}~(player), \texttt{O}~(box), \texttt{X}~(goal), \texttt{*}~(box on goal), and \texttt{+}~(player on goal), with row breaks stripped.
\begin{lstlisting}
{"g":"######@.X##.O.##...######","moves":3}
\end{lstlisting}

\textbf{Bloxorz.}
The grid encodes the block state (\texttt{S}~standing, \texttt{=}~or~\texttt{-}~lying along X, \texttt{H}~or~\texttt{\textbar}~lying along Y), along with the block position and orientation, the goal location, the number of broken fragile tiles, and the Manhattan distance from the block to the goal.
\begin{lstlisting}
{"g":".....S.......G......","b":[1,1,"S"],
 "goal":[3,3],"broken":0,"distance":4,"moves":2}
\end{lstlisting}

\textbf{Rush Hour.}
The grid encodes vehicle letters at their occupied cells.
Additional fields record the exit position, the number of vehicles, and a list of vehicle letters blocking car~A's path to the exit.
\begin{lstlisting}
{"g":"......B.....AA.B..C.CC...........",
 "exit_row":2,"exit_col":5,"n_vehicles":3,
 "blocks_A":["B","C"],"moves":1}
\end{lstlisting}

\textbf{Anagram.}
The schema records the current letter arrangement, the sorted letters, the string length, whether an identification has been attempted, the number of valid anagram solutions, and per-letter frequency counts.
\begin{lstlisting}
{"letters":"cat","sorted":"act","len":3,
 "id":false,"n_valid":1,
 "letter_counts":"a:1 c:1 t:1","moves":0}
\end{lstlisting}

\textbf{Tangram.}
The canvas grid uses digits to mark cells covered by placed pieces (the last character of the piece identifier), \texttt{T} for uncovered target cells, and \texttt{.} for empty non-target cells.
An array records the identifier and rotation of each unplaced piece.
\begin{lstlisting}
{"c":"00T..01T..11T..TTT..",
 "q_rot":[["P2",90]],"moves":2}
\end{lstlisting}

\textbf{Jigsaw.}
A flat array records piece identifiers at each grid cell (\texttt{null} for empty cells), along with the rotation of every piece, a list of unplaced piece identifiers, and the count of placed pieces.
\begin{lstlisting}
{"g":[0,1,null,3],"r":[0,0,90,0],
 "unplaced":[2],"n_placed":3,"moves":4}
\end{lstlisting}

\textbf{Character Recognition.}
The schema records the rotation of each piece and its slot assignment (\texttt{null} if unplaced), along with the number of placed pieces.
\begin{lstlisting}
{"r":[0,90,0,0],"p":[2,0,null,null],
 "n_placed":2,"moves":1}
\end{lstlisting}

\textbf{3D Mental Rotation.}
The schema records the current rotation angles (\texttt{rx}, \texttt{ry}, \texttt{rz}), the target angles (\texttt{tx}, \texttt{ty}, \texttt{tz}), the signed shortest angular delta per axis (\texttt{dx}, \texttt{dy}, \texttt{dz}), the rotation increment, and the number of voxels in the polycube.
\begin{lstlisting}
{"rx":0,"ry":0,"rz":0,"tx":90,"ty":180,"tz":0,
 "dx":90,"dy":180,"dz":0,
 "inc":15,"n_voxels":7,"moves":0}
\end{lstlisting}

\textbf{2D Mental Rotation.}
The schema records the current rotation, the target rotation, the signed shortest angular delta, the rotation increment, and the number of cells in the polyomino.
\begin{lstlisting}
{"r":45,"target":135,"delta":90,
 "inc":15,"n_cells":5,"moves":3}
\end{lstlisting}

\clearpage
\section{BAGEL Baseline Details}
\label{sec:bagel_details}

This appendix describes our adaptation of BAGEL-7B-MoT~\cite{deng2025bagel} to the puzzle suite via behavioral cloning and our analysis of why it failed to converge when being trained with its visual head.

\textbf{Setup.}
BAGEL is a 7B-parameter Mixture-of-Transformers model that natively handles both image understanding (SigLIP ViT + connector) and image generation (VAE + flow-matching head) within a single architecture.
We adapt it to our puzzle suite by constructing interleaved trajectories: the user turn contains ViT-encoded patches of the initial image plus the puzzle prompt, and the assistant turn alternates between blocks of 196 VAE latent tokens, encoding the world state after each action as a $14 \times 14$ continuous latent grid, with discrete action tokens enclosed in custom \texttt{<action>}/\texttt{</action>} delimiters.
Images are rendered at $128$\,px by the same game engine used for all other methods and encoded by BAGEL's frozen VAE.

\textbf{Loss design.}
Each trajectory position carries one of two losses: (i)~standard shifted next-token cross-entropy on the action-token spans, or (ii)~flow-matching MSE on the VAE velocity-field head at the 196 latent positions of each generated frame, trained under a Rectified-Flow schedule.
The total loss is $\mathcal{L} = \alpha\,\text{mean}(\text{CE}) + \beta\,\text{mean}(\text{MSE})$.
Because the MSE averages over ${\sim}196$ tokens per frame while the CE averages over ${\sim}5$--$10$ action tokens per step, the per-episode gradient contribution is strongly tilted toward reconstruction even at nominally equal weights.
We train with LoRA adapters (rank 16) on all attention and MLP layers including the vision connector and the MoT generation head, using AdamW with bf16 and FlashAttention-2.

\textbf{Training pathology.}
The naive approach feeds the full interleaved sequence causally. Each action attends to every prior VAE-latent block and every prior action.
This collapsed training for two compounding reasons.
First, effective context explodes: with 196 VAE tokens per frame, a medium-difficulty episode consumes thousands of context tokens, and self-attention becomes dominated by cross-frame latent-to-latent interactions that carry minimal task signal.
Second, error accumulates at evaluation: under teacher forcing the model sees ground-truth VAE latents, but at rollout it sees its own noisy generations.
Because each action conditions on the full visual history, errors at frame~$t$ corrupt the context for all subsequent steps.
In practice, multi-step solve rate stagnated near zero despite reasonable single-step accuracy -- a classic sign of overfitting to the teacher-forced visual stream rather than learning a coherent policy. We modified BAGEL to just generate the visual state without keeping it in the attention KV-cache or attend only to one past state. This improved the training but still fell well short of the baseline.

\clearpage
\section{Additional Results}

\subsection{Small VLM results}
\begin{figure}[p]
\centering
\captionsetup[subfigure]{skip=2pt,font=scriptsize,justification=centering}
\pgfplotsset{
    gameplot/.style={
        scale only axis, width=1.75cm, height=1.8cm,
        ytick=\empty,
        ymin=0, ymax=5,
        xmajorgrids,
        major grid style={dashed, gray!30},
        xticklabel style={font=\tiny},
        every axis/.append style={thin},
        axis x line*=bottom,
        axis y line*=left,
    },
}
\newcommand{\gameplot}[6]{%
\begin{tikzpicture}
\pgfmathtruncatemacro{\xmaxpad}{#6+12}%
\begin{axis}[gameplot, xmin=#5, xmax=\xmaxpad, xtick={#5,#6}]
\pgfmathsetmacro{\va}{max(#1,#5)}
\pgfmathsetmacro{\vb}{max(#2,#5)}
\pgfmathsetmacro{\vc}{max(#3,#5)}
\pgfmathsetmacro{\vd}{max(#4,#5)}
\fill[flatConcrete] (axis cs:#5,3.6) rectangle (axis cs:\va,4.4);
\node[right, font=\tiny, inner sep=1pt] at (axis cs:\va,4) {#1};
\fill[flatGreenSea] (axis cs:#5,2.6) rectangle (axis cs:\vb,3.4);
\node[right, font=\tiny, inner sep=1pt] at (axis cs:\vb,3) {#2};
\fill[flatOrange] (axis cs:#5,1.6) rectangle (axis cs:\vc,2.4);
\node[right, font=\tiny, inner sep=1pt] at (axis cs:\vc,2) {#3};
\fill[flatPomegranate] (axis cs:#5,0.6) rectangle (axis cs:\vd,1.4);
\node[right, font=\tiny, inner sep=1pt] at (axis cs:\vd,1) {#4};
\end{axis}
\end{tikzpicture}%
}%
\begin{subfigure}[t]{0.135\textwidth}\centering
\gameplot{100.0}{100.0}{100.0}{100.0}{70}{100}
\caption{2D Shape Match.}
\end{subfigure}\hfill
\begin{subfigure}[t]{0.135\textwidth}\centering
\gameplot{100.0}{100.0}{100.0}{100.0}{70}{100}
\caption{2D Mental Rot.}
\end{subfigure}\hfill
\begin{subfigure}[t]{0.135\textwidth}\centering
\gameplot{46.0}{73.5}{71.0}{75.5}{40}{80}
\caption{3D Shape Match.}
\end{subfigure}\hfill
\begin{subfigure}[t]{0.135\textwidth}\centering
\gameplot{87.0}{90.5}{89.0}{95.5}{80}{100}
\caption{3D Mental Rot.}
\end{subfigure}\hfill
\begin{subfigure}[t]{0.135\textwidth}\centering
\gameplot{97.0}{97.5}{98.5}{96.5}{90}{100}
\caption{Bloxorz}
\end{subfigure}\hfill
\begin{subfigure}[t]{0.135\textwidth}\centering
\gameplot{95.5}{90.5}{99.0}{96.5}{80}{100}
\caption{Rush Hour}
\end{subfigure}\hfill
\begin{subfigure}[t]{0.135\textwidth}\
\end{subfigure}\\[0.088in]
\begin{subfigure}[t]{0.135\textwidth}\centering
\gameplot{100.0}{100.0}{100.0}{100.0}{70}{100}
\caption{Char.\ Recog.}
\end{subfigure}\hfill
\begin{subfigure}[t]{0.135\textwidth}\centering
\gameplot{62.0}{62.5}{85.0}{63.0}{60}{100}
\caption{Jigsaw}
\end{subfigure}\hfill
\begin{subfigure}[t]{0.135\textwidth}\centering
\gameplot{56.0}{52.5}{64.0}{64.0}{40}{70}
\caption{Tangram}
\end{subfigure}\hfill
\begin{subfigure}[t]{0.135\textwidth}\centering
\gameplot{68.5}{67.0}{74.0}{69.5}{60}{80}
\caption{Tangram (y/n)}
\end{subfigure}\hfill
\begin{subfigure}[t]{0.135\textwidth}\centering
\gameplot{95.5}{89.0}{98.0}{94.5}{80}{100}
\caption{Sokoban}
\end{subfigure}\hfill
\begin{subfigure}[t]{0.135\textwidth}\centering
\gameplot{100.0}{99.5}{100.0}{100.0}{95}{100}
\caption{Anagram}
\end{subfigure}\hfill
\begin{subfigure}[t]{0.135\textwidth}\centering
\gameplot{84.0}{85.2}{89.9}{87.9}{70}{100}
\caption{\textbf{Average}}
\end{subfigure}\\[4pt]
\begin{tikzpicture}[
lgsq/.style={minimum width=8pt, minimum height=6pt, inner sep=0pt, yshift=2pt},
lglbl/.style={font=\footnotesize, inner sep=1pt, anchor=base west,
              text height=1.6ex, text depth=.3ex},
]
\matrix[column sep=4pt, nodes={anchor=base}, ampersand replacement=\&]{
\node[lgsq, fill=flatConcrete]{};    \& \node[lglbl]{Action Supervision};                          \&
\node[lgsq, fill=flatGreenSea]{};    \& \node[lglbl]{Visual Supervision};                  \&
\node[lgsq, fill=flatOrange]{};      \& \node[lglbl]{Visual Tokens $64{\times}64$};    \&
\node[lgsq, fill=flatPomegranate]{}; \& \node[lglbl]{Visual Tokens $128{\times}128$};  \\
};
\end{tikzpicture}
\caption{\textbf{Solve rates on visual reasoning tasks (Qwen~3.5 2B).} Each chart shows the results of the Base Qwen3.5, No CoT and various forms of supervison. Visual Tokens $64{\times}64$ leads the average (89.9\%); the largest gains over No-CoT come on 3D Shape Match (+27\,pts for Visual Superv.) and Jigsaw (+23\,pts for Visual Tokens $64{\times}64$).}
\label{fig:overview_2b}
\end{figure}
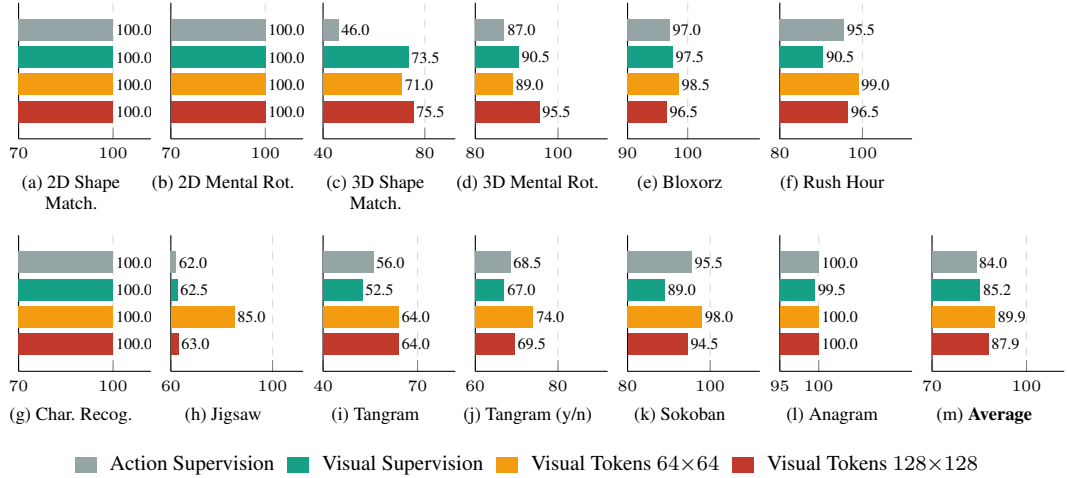

In \cref{fig:overview_0_8b,fig:overview_2b}, we show the puzzle solve rates of fine-tuned Qwen3.5 0.8B and 2B, respectively. The action-only method achieves solve rates of 82\% and 84\% in the two cases, and using visual CoT further enhances the solve rates on the most complex puzzles. This shows that our primary findings on the Qwen3.5 9B model hold for smaller models as well.

\begin{figure}[p]
\centering
\captionsetup[subfigure]{skip=2pt,font=scriptsize,justification=centering}
\pgfplotsset{
    gameplot/.style={
        scale only axis, width=1.75cm, height=1.8cm,
        ytick=\empty,
        ymin=0, ymax=5,
        xmajorgrids,
        major grid style={dashed, gray!30},
        xticklabel style={font=\tiny},
        every axis/.append style={thin},
        axis x line*=bottom,
        axis y line*=left,
    },
}
\newcommand{\gameplot}[6]{%
\begin{tikzpicture}
\pgfmathtruncatemacro{\xmaxpad}{#6+12}%
\begin{axis}[gameplot, xmin=#5, xmax=\xmaxpad, xtick={#5,#6}]
\pgfmathsetmacro{\va}{max(#1,#5)}
\pgfmathsetmacro{\vb}{max(#2,#5)}
\pgfmathsetmacro{\vc}{max(#3,#5)}
\pgfmathsetmacro{\vd}{max(#4,#5)}
\fill[flatConcrete] (axis cs:#5,3.6) rectangle (axis cs:\va,4.4);
\node[right, font=\tiny, inner sep=1pt] at (axis cs:\va,4) {#1};
\fill[flatGreenSea] (axis cs:#5,2.6) rectangle (axis cs:\vb,3.4);
\node[right, font=\tiny, inner sep=1pt] at (axis cs:\vb,3) {#2};
\fill[flatOrange] (axis cs:#5,1.6) rectangle (axis cs:\vc,2.4);
\node[right, font=\tiny, inner sep=1pt] at (axis cs:\vc,2) {#3};
\fill[flatPomegranate] (axis cs:#5,0.6) rectangle (axis cs:\vd,1.4);
\node[right, font=\tiny, inner sep=1pt] at (axis cs:\vd,1) {#4};
\end{axis}
\end{tikzpicture}%
}%
\begin{subfigure}[t]{0.135\textwidth}\centering
\gameplot{100.0}{100.0}{100.0}{100.0}{70}{100}
\caption{2D Shape Match.}
\end{subfigure}\hfill
\begin{subfigure}[t]{0.135\textwidth}\centering
\gameplot{100.0}{100.0}{100.0}{100.0}{70}{100}
\caption{2D Mental Rot.}
\end{subfigure}\hfill
\begin{subfigure}[t]{0.135\textwidth}\centering
\gameplot{48.0}{57.5}{77.0}{67.5}{40}{80}
\caption{3D Shape Match.}
\end{subfigure}\hfill
\begin{subfigure}[t]{0.135\textwidth}\centering
\gameplot{87.5}{85.0}{92.0}{96.0}{80}{100}
\caption{3D Mental Rot.}
\end{subfigure}\hfill
\begin{subfigure}[t]{0.135\textwidth}\centering
\gameplot{96.0}{95.0}{97.5}{94.5}{90}{100}
\caption{Bloxorz}
\end{subfigure}\hfill
\begin{subfigure}[t]{0.135\textwidth}\centering
\gameplot{90.5}{85.0}{96.0}{96.5}{80}{100}
\caption{Rush Hour}
\end{subfigure}\hfill
\begin{subfigure}[t]{0.135\textwidth}\
\end{subfigure}\\[0.088in]
\begin{subfigure}[t]{0.135\textwidth}\centering
\gameplot{100.0}{100.0}{100.0}{100.0}{70}{100}
\caption{Char.\ Recog.}
\end{subfigure}\hfill
\begin{subfigure}[t]{0.135\textwidth}\centering
\gameplot{61.5}{68.5}{61.5}{92.5}{60}{100}
\caption{Jigsaw}
\end{subfigure}\hfill
\begin{subfigure}[t]{0.135\textwidth}\centering
\gameplot{47.0}{42.5}{55.5}{56.5}{40}{70}
\caption{Tangram}
\end{subfigure}\hfill
\begin{subfigure}[t]{0.135\textwidth}\centering
\gameplot{64.0}{66.5}{68.0}{66.0}{60}{80}
\caption{Tangram (y/n)}
\end{subfigure}\hfill
\begin{subfigure}[t]{0.135\textwidth}\centering
\gameplot{89.5}{87.0}{93.5}{84.5}{80}{100}
\caption{Sokoban}
\end{subfigure}\hfill
\begin{subfigure}[t]{0.135\textwidth}\centering
\gameplot{99.5}{99.5}{100.0}{100.0}{95}{100}
\caption{Anagram}
\end{subfigure}\hfill
\begin{subfigure}[t]{0.135\textwidth}\centering
\gameplot{82.0}{82.2}{86.8}{87.8}{70}{100}
\caption{\textbf{Average}}
\end{subfigure}\\[4pt]
\begin{tikzpicture}[
    lgsq/.style={minimum width=8pt, minimum height=6pt, inner sep=0pt, yshift=2pt},
    lglbl/.style={font=\footnotesize, inner sep=1pt, anchor=base west,
                  text height=1.6ex, text depth=.3ex},
  ]
  \matrix[column sep=4pt, nodes={anchor=base}, ampersand replacement=\&]{
    \node[lgsq, fill=flatConcrete]{};    \& \node[lglbl]{Action Supervision};                          \&
    \node[lgsq, fill=flatGreenSea]{};    \& \node[lglbl]{Visual Supervision};                  \&
    \node[lgsq, fill=flatOrange]{};      \& \node[lglbl]{Visual Tokens $64{\times}64$};    \&
    \node[lgsq, fill=flatPomegranate]{}; \& \node[lglbl]{Visual Tokens $128{\times}128$};  \\
  };
  \end{tikzpicture}
\caption{\textbf{Solve rates on visual reasoning tasks (Qwen~3.5 0.8B).} Each chart shows the results of the Base Qwen3.5, action supervision and various forms of visual supervison. Visual tokens-based CoT improves over action supervision on every non-saturated game; the largest gains come on 3D Shape Match (+29\,pts for Visual Tokens $64{\times}64$) and Jigsaw (+31\,pts for Visual Tokens $128{\times}128$).}
\label{fig:overview_0_8b}
\end{figure}
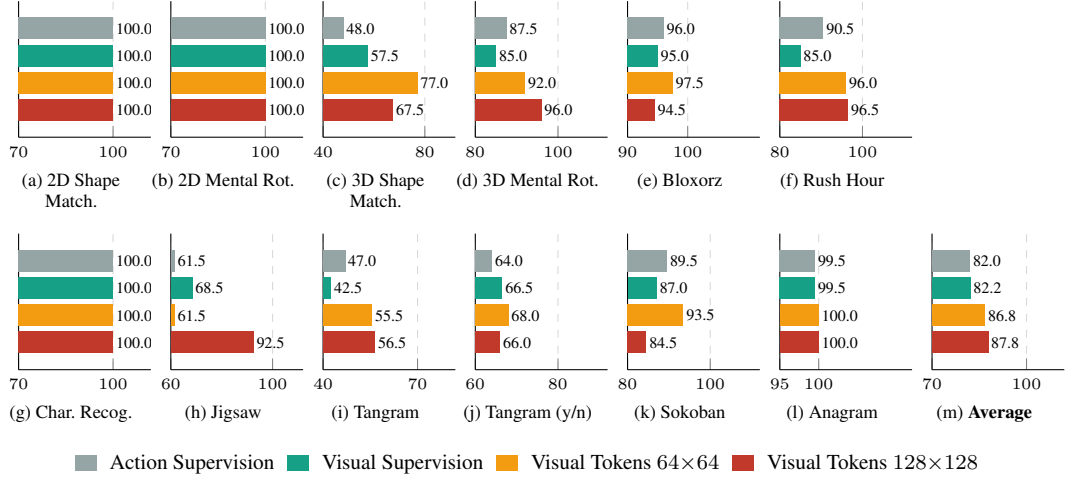

\subsection{Comparing LlamaGen vs. FSQ tokenizers for visual token generation}
\label{sec:llamagen_vs_fsq}
\begin{figure}
\centering
\captionsetup[subfigure]{skip=2pt,font=scriptsize,justification=centering}
\pgfplotsset{
    gameplot/.style={
        scale only axis, width=1.75cm, height=1.8cm,
        ytick=\empty,
        ymin=0, ymax=5,
        xmajorgrids,
        major grid style={dashed, gray!30},
        xticklabel style={font=\tiny},
        every axis/.append style={thin},
        axis x line*=bottom,
        axis y line*=left,
    },
}
\newcommand{\gameplot}[6]{%
\begin{tikzpicture}
\pgfmathtruncatemacro{\xmaxpad}{#6+12}%
\begin{axis}[gameplot, xmin=#5, xmax=\xmaxpad, xtick={#5,#6}]
\pgfmathsetmacro{\va}{max(#1,#5)}
\pgfmathsetmacro{\vb}{max(#2,#5)}
\pgfmathsetmacro{\vc}{max(#3,#5)}
\pgfmathsetmacro{\vd}{max(#4,#5)}
\fill[flatOrange] (axis cs:#5,3.6) rectangle (axis cs:\va,4.4);
\node[right, font=\tiny, inner sep=1pt] at (axis cs:\va,4) {#1};
\fill[flatPomegranate] (axis cs:#5,2.6) rectangle (axis cs:\vb,3.4);
\node[right, font=\tiny, inner sep=1pt] at (axis cs:\vb,3) {#2};
\fill[flatPeterRiver] (axis cs:#5,1.6) rectangle (axis cs:\vc,2.4);
\node[right, font=\tiny, inner sep=1pt] at (axis cs:\vc,2) {#3};
\fill[flatBelizeHole] (axis cs:#5,0.6) rectangle (axis cs:\vd,1.4);
\node[right, font=\tiny, inner sep=1pt] at (axis cs:\vd,1) {#4};
\end{axis}
\end{tikzpicture}%
}%
\begin{subfigure}[t]{0.135\textwidth}\centering
\gameplot{100.0}{100.0}{100.0}{100.0}{70}{100}
\caption{2D Shape Match.}
\end{subfigure}\hfill
\begin{subfigure}[t]{0.135\textwidth}\centering
\gameplot{100.0}{100.0}{100.0}{100.0}{70}{100}
\caption{2D Mental Rot.}
\end{subfigure}\hfill
\begin{subfigure}[t]{0.135\textwidth}\centering
\gameplot{91.0}{97.5}{86.0}{96.5}{70}{100}
\caption{3D Shape Match.}
\end{subfigure}\hfill
\begin{subfigure}[t]{0.135\textwidth}\centering
\gameplot{70.5}{77.5}{66.5}{66.0}{50}{80}
\caption{3D Mental Rot.}
\end{subfigure}\hfill
\begin{subfigure}[t]{0.135\textwidth}\centering
\gameplot{97.5}{94.5}{97.0}{94.5}{70}{100}
\caption{Bloxorz}
\end{subfigure}\hfill
\begin{subfigure}[t]{0.135\textwidth}\centering
\gameplot{95.0}{97.0}{93.5}{92.5}{70}{100}
\caption{Rush Hour}
\end{subfigure}\hfill
\begin{subfigure}[t]{0.135\textwidth}\
\end{subfigure}\\[0.088in]
\begin{subfigure}[t]{0.135\textwidth}\centering
\gameplot{100.0}{100.0}{99.5}{100.0}{70}{100}
\caption{Char.\ Recog.}
\end{subfigure}\hfill
\begin{subfigure}[t]{0.135\textwidth}\centering
\gameplot{84.5}{91.5}{92.0}{93.0}{60}{100}
\caption{Jigsaw}
\end{subfigure}\hfill
\begin{subfigure}[t]{0.135\textwidth}\centering
\gameplot{59.0}{55.0}{62.5}{53.0}{40}{70}
\caption{Tangram}
\end{subfigure}\hfill
\begin{subfigure}[t]{0.135\textwidth}\centering
\gameplot{75.0}{72.0}{68.5}{70.0}{50}{80}
\caption{Tangram (y/n)}
\end{subfigure}\hfill
\begin{subfigure}[t]{0.135\textwidth}\centering
\gameplot{95.0}{89.5}{97.0}{85.0}{70}{100}
\caption{Sokoban}
\end{subfigure}\hfill
\begin{subfigure}[t]{0.135\textwidth}\centering
\gameplot{100.0}{100.0}{97.5}{99.0}{70}{100}
\caption{Anagram}
\end{subfigure}\hfill
\begin{subfigure}[t]{0.135\textwidth}\centering
\gameplot{89.0}{89.5}{88.3}{87.5}{70}{100}
\caption{\textbf{Average}}
\end{subfigure}\\[4pt]
\begin{tikzpicture}
\newcommand{\legenditem}[4]{\fill[#1] (#2,#3) rectangle ++(0.3,0.2); \node[right, font=\footnotesize] at (#2+0.35,#3+0.1) {#4};}
\legenditem{flatOrange}{0}{0.5}{Visual Tokens $64{\times}64$ (llamagen)}
\legenditem{flatPomegranate}{5.2}{0.5}{Visual Tokens $128{\times}128$ (llamagen)}
\legenditem{flatPeterRiver}{0}{0}{Visual Tokens $64{\times}64$ (FSQ)}
\legenditem{flatBelizeHole}{5.2}{0}{Visual Tokens $128{\times}128$ (FSQ)}
\end{tikzpicture}
\caption{\textbf{Comparing Llamagen vs. FSQ tokenizers for visual token prediction.} We observe that Llamagen tokens are on average better for autoregressive generation than the FSQ tokens. Therefore, we use it as our de-facto approach for autoregressive generation.}
\label{fig:llamagen_vs_fsq}
\end{figure}
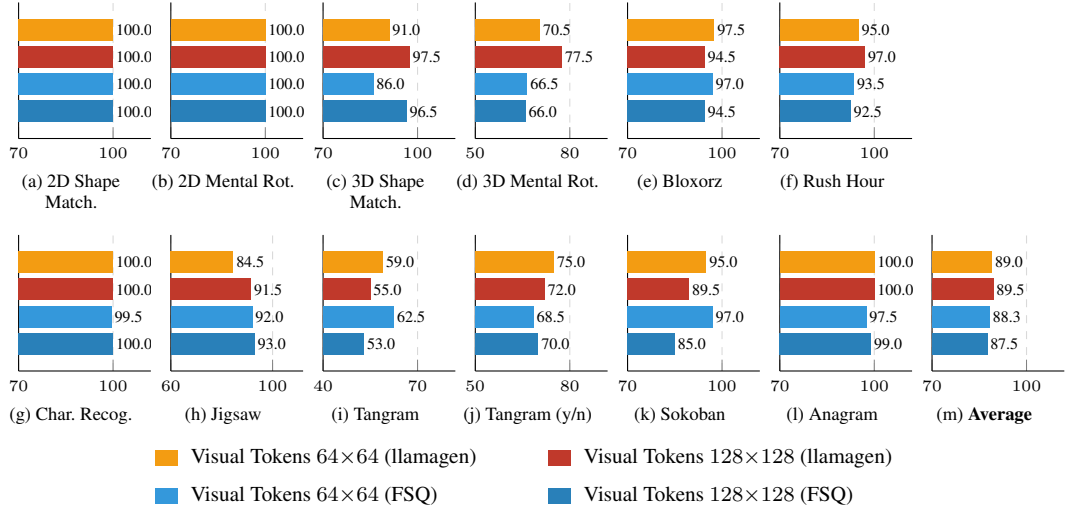

In~\cref{fig:llamagen_vs_fsq}, we compare the impact of using LlamaGen and FSQ tokens for visual token-based CoT. On average, we observe that LlamaGen was better than FSQ by 1\% on our puzzle suite. Therefore, we use LlamaGen as the default tokenizer for the visual token experiments. 

\subsection{Anagram Analysis}
\label{sec:anagram}

\begin{figure}[p]
    \centering
    \begin{tabular}{ccc}
        \includegraphics[width=0.2\linewidth]{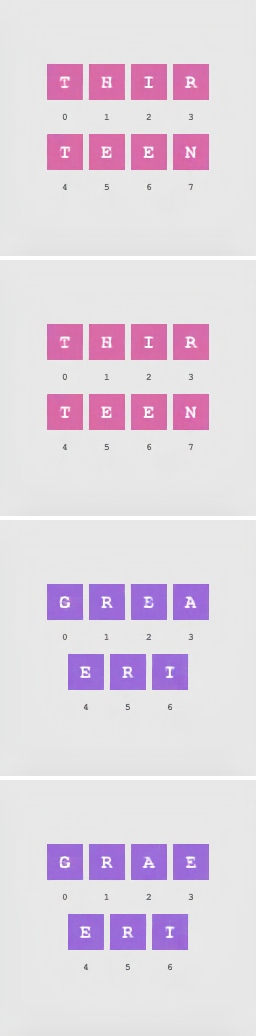} &
        \includegraphics[width=0.2\linewidth]{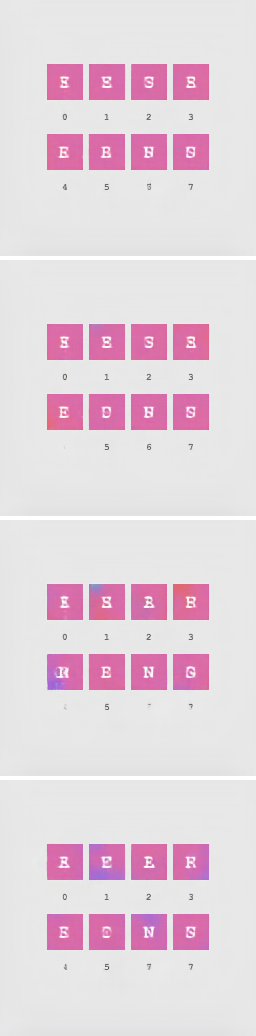} &
        \includegraphics[width=0.2\linewidth]{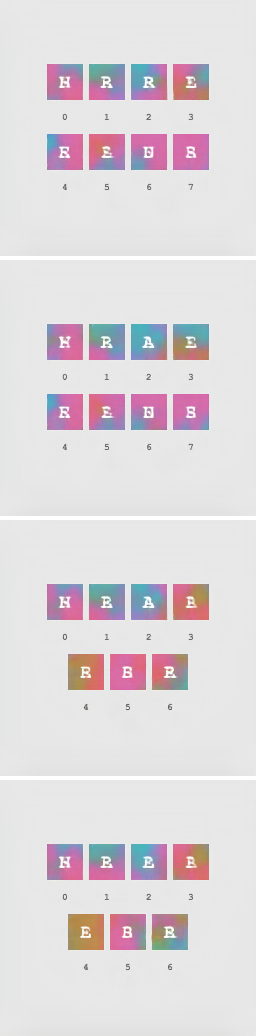} \\
        (a) Ground truth & (b) Base Qwen & (c) Action Only
    \end{tabular}
    \caption{\textbf{Anagram reconstruction comparison.} (a)~The ground-truth visual state. (b)~Reconstruction from the Base Qwen model, which produces nearly uniform reconstructions that vary little across inputs, achieving low reconstruction error by approximating a mean image that is uninformative. (c)~The detached trained visual head for action-supervised model reconstructs images with incorrect colors yielding higher reconstruction error. Yet, it preserves the spatial arrangement and letter identities of the actual input, producing more semantically meaningful representations.}
    \label{fig:anagram_recon}
\end{figure}

As noted in~\cref{fig:psnr_vs_solve}, anagram is the main exception to the general trend where improved reconstruction quality correlates with higher solve rates. However, this apparent discrepancy is explained by examining the nature of the reconstructions themselves. \cref{fig:anagram_recon} reveals that the base model achieves lower reconstruction error by producing nearly uniform outputs that vary little across different inputs, approximating a mean image. Because anagram images share the same general structure -- colored tiles on a grey background -- this mean approximation achieves low pixel-wise error but encodes no task-relevant information about letter positions or identities. In contrast, the action-supervised model produces reconstructions with incorrect tile colors (yielding higher MSE) but faithfully preserves the spatial layout and letter content of each specific input. Thus, the action-supervised model's representations are more informative for solving the task despite their higher reconstruction error, confirming that the correlation between mental image quality and solve rate holds for anagram as well when quality is measured by semantic fidelity rather than pixel-wise distance.

\clearpage
\section{Qualitative Rollouts}

In the remainder of the pages, we show qualitative rollouts from different models and different games.

\newcommand{\modelActionOnly}{Action-only Supervision}
\newcommand{\modelVH}{Action and Visual Supervision}
\newcommand{\modelVQVAE}{Visual Tokens}

\newcommand{\rolloutfigBase}[6]{%
    \begin{figure}[p]
      \centering
      \includegraphics[width=\linewidth,height=0.85\textheight,keepaspectratio]{#1/#2_figure.pdf}
      \caption{Autoregressive rollouts for \textbf{#3} on \textbf{#6}.
        See also: \hyperref[fig:#2:#4]{#4 (Fig.~\ref*{fig:#2:#4})} and \hyperref[fig:#2:#5]{#5 (Fig.~\ref*{fig:#2:#5})}.}
      \label{fig:#2:#3}
    \end{figure}%
  }

\newcommand{\rolloutfigActionOnly}[2]{%
\rolloutfigBase{video_actiononly_vh}{#1}{\modelActionOnly}{\modelVH}{\modelVQVAE}{#2}%
}
\newcommand{\rolloutfigVH}[2]{%
\rolloutfigBase{video_vh}{#1}{\modelVH}{\modelActionOnly}{\modelVQVAE}{#2}%
}
\newcommand{\rolloutfigVQVAE}[2]{%
\rolloutfigBase{video_vqvae}{#1}{\modelVQVAE}{\modelActionOnly}{\modelVH}{#2}%
}

\newcommand{\eachGame}[1]{%
#1{anagram}{Anagram}%
#1{bloxorz}{Bloxorz}%
#1{character_recognition}{Character Recognition}%
#1{jigsaw}{Jigsaw}%
#1{mental_rotation}{Shape Matching (3D)}%
#1{mental_rotation_certificate}{Mental Rotation (3D) (yes/no certificate)}%
#1{mental_rotation_2d}{Shape Matching (2D)}%
#1{mental_rotation_2d_certificate}{Mental Rotation (2D) (yes/no certificate)}%
#1{rush_hour}{Rush Hour}%
#1{sokoban}{Sokoban}%
#1{tangram}{Tangram}%
#1{tangram_certificate}{Tangram (yes/no certificate)}%
}

\newcommand{\rolloutTocItem}[2]{%
\item[#2]\hfill\\
  Fig.~\ref*{fig:#1:\modelActionOnly}: \hyperref[fig:#1:\modelActionOnly]{\modelActionOnly}\\
  Fig.~\ref*{fig:#1:\modelVH}: \hyperref[fig:#1:\modelVH]{\modelVH}\\
  Fig.~\ref*{fig:#1:\modelVQVAE}: \hyperref[fig:#1:\modelVQVAE]{\modelVQVAE}%
}

\begin{description}
\eachGame{\rolloutTocItem}
\end{description}

\eachGame{\rolloutfigActionOnly}
\eachGame{\rolloutfigVH}
\eachGame{\rolloutfigVQVAE}

\end{document}